\documentclass[journal]{IEEEtai}

\usepackage[colorlinks,urlcolor=blue,linkcolor=blue,citecolor=blue]{hyperref}
\usepackage{color,array}
\usepackage{graphicx}
\usepackage{amsmath,amsfonts}
\usepackage{algorithmic}
\usepackage{verbatim}
\usepackage{algorithm,algorithmic}
\usepackage{mathtools}
\usepackage{subfigure}
\usepackage{rotating}
\usepackage{hyperref}
\usepackage{adjustbox}
\usepackage{booktabs,makecell}
\newcommand{\quotes}[1]{``#1''}

\begin{document}

\title{Challenges and Limitations of Generative AI in Synthesizing Wearable Sensor Data}
\author{Flavio Di Martino, and Franca Delmastro
\thanks{Flavio Di Martino and Franca Delmastro are with the Institute for Informatics and Telematics of the Nationa Research Council (IIT-CNR) of Italy, Pisa, Italy (IT) (e-mail: name.surname@iit.cnr.it).}}

\markboth{Journal of IEEE Transactions on Artificial Intelligence, Vol. 00, No. 0, Month 2020}
{First A. Author \MakeLowercase{\textit{et al.}}: Bare Demo of IEEEtai.cls for IEEE Journals of IEEE Transactions on Artificial Intelligence}

\maketitle

\begin{abstract}
The widespread adoption of wearable sensors has the potential to provide massive and heterogeneous time series data, driving the use of Artificial Intelligence in human sensing applications.
However, data collection remains limited due to stringent ethical regulations, privacy concerns, and other constraints, hindering progress in the field. Synthetic data generation, particularly through Generative Adversarial Networks and Diffusion Models, has emerged as a promising solution to mitigate both data scarcity and privacy issues. 
However, these models are often limited to narrow operational scenarios, such as short-term and unimodal signal patterns. To address this gap, we present a systematic evaluation of state-of-the-art generative models for time series data, explicitly assessing their performance in challenging scenarios such as stress and emotion recognition.
Our study examines the extent to which these models can jointly handle multi-modality, capture long-range dependencies, and support conditional generation—core requirements for real-world wearable sensor data generation. To enable a fair and rigorous comparison, we also introduce an evaluation framework that evaluates both the intrinsic fidelity of the generated data and their utility in downstream predictive tasks.
Our findings reveal critical limitations in the existing approaches, particularly in maintaining cross-modal consistency, preserving temporal coherence, and ensuring robust performance in train-on-synthetic, test-on-real, and data augmentation scenarios. Finally, we present our future research directions to enhance synthetic time series generation and improve the applicability of generative models in the wearable computing domain.
\end{abstract}

\begin{IEEEImpStatement}
The scarcity and imbalance of real-world wearable sensor data remain major barriers to AI-powered human sensing applications. This study presents a comprehensive evaluation of state-of-the-art generative frameworks for time series data, introducing a challenging and realistic benchmark that jointly addresses multimodal, long-sequence, and conditional generation—dimensions rarely explored together in previous work.
Our experiments reveal critical shortcomings, including substantial cross-modal quality gaps and limited effectiveness in downstream tasks, with performance declines exceeding $10\%$ in train-on-synthetic, test-on-real scenarios and only modest improvements ($2$–$3\%$) from data augmentation.
These findings underscore the practical limitations of current generative approaches while providing guidance for advancing sensor data generation across domains as wearable computing, behavioral analytics, and human-AI interaction.
\end{IEEEImpStatement}


\section{Introduction}
\label{sec:introduction}

\IEEEPARstart{R}{ecent} breakthroughs in Artificial Intelligence (AI) have been largely fueled by the availability of large-scale datasets, such as image repositories and text corpora.
Meanwhile, the proliferation of wearables, smartphones, and Internet of Things (IoT) devices has enabled continuous and unobtrusive collection of rich time series (TS) data, driving innovation across several human sensing applications such as mobile health (mHealth) \cite{triantafyllidis2022deep} and human activity recognition (HAR) \cite{yu2024human}.
Despite these advancements, the collection and utilization of such data remain significantly constrained by stringent privacy and ethical regulations, low user compliance, scarce annotations, and other logistical and technical challenges.
As a result, AI research often depends on small, private, and fragmented datasets, hindering the development and validation of robust tools. Overcoming these barriers is essential for unlocking the full potential of AI in human sensing and enabling its seamless deployment in real-world systems. In this context, synthetic data generation emerges as a promising solution for addressing these limitations. Synthetic data can facilitate the creation and sharing of realistic digital twins of private datasets, effectively mitigating privacy concerns while preserving the statistical properties of real-world data. Additionally, synthetic data can enhance predictive modeling by replacing or augmenting real-world samples, offering an alternative to paradigms such as transfer learning and few-shot learning in low-data regimes. Consequently, generative models have attracted considerable attention as promising tools for advancing human sensing applications.
\\
In recent years, Generative Adversarial Networks (GAN) \cite{goodfellow2014generative} and diffusion models \cite{sohl2015deep} have achieved remarkable success in computer vision (CV) \cite{radford2015unsupervised,zhu2017unpaired,dhariwal2021diffusion, rombach2022high}. Extending these techniques to TS generation, however, presents unique challenges. Unlike images with fixed spatial structures, TS data are inherently sequential and characterized by temporal dependencies that must be preserved to ensure overall coherence. Moreover, generating heterogeneous wearable sensor data, such as physiological and behavioral signals, introduces additional complexities that must be carefully managed to ensure real-world applicability.
One critical challenge lies in the effective fusion of multimodal data. Prior studies have shown that integrating multiple sensing modalities can enhance predictive performance across diverse applications \cite{katada2022effects,junaid2023explainable,gao2024explainable}.
Consequently, generative models must be capable of jointly synthesizing multiple signals while maintaining cross-modal coherence. Additionally, the ability to generate long sequences is essential for applications involving high-rate biosignals or extended monitoring windows, as insufficient sequence length can compromise downstream inference.
Conditional generation further enhances model flexibility by enabling a single training run to synthesize data across different categories, while incorporating contextual metadata to guide and refine the generative process.
Eventually, evaluating synthetic TS is an open challenge because, unlike images, there are no universally accepted benchmarking standards. Thus, developing standardized and objective evaluation frameworks is essential to enhance the reliability of generative TS models and advance AI applications in human sensing.
\\
In recent years, TS generative models have shown promising results, yet generally under strict operational constraints. Most models are designed for single, specific signals, with limited scalability to multi-axis or multi-channel data
from the same modality. Additionally, common benchmarks for multivariate TS (MTS) generation often rely on overly simplistic datasets (e.g., \textit{sines}, \textit{stocks}, and \textit{energy}), which are also not representative of mobile sensor data.
Many studies also concentrate on datasets and applications dominated by short-term patterns, where models can perform well without requiring extensive memory or complex mechanisms to capture long-range dependencies (LRD).
Furthermore, generative models are frequently trained in an unconditional fashion, relying on a consistent source distribution. Conditional generation, while promising, remains relatively unexplored and is often limited to basic class labels, restricting customization at both cohort and subject level.
\\
Given these limitations, we argue that current state-of-the-art (SoTA) TS generative models are not readily applicable to wearable sensor data generation. Therefore, to assess the feasibility and performance of synthetic data generation in this scenario, this study targets a more challenging task: multimodal, long-range, and conditional TS generation.
We focus on a comprehensive analysis of SoTA solutions specifically tailored to TS synthesis, including the most relevant GAN and the latest diffusion models. Our study evaluates these models using real-world wearable datasets and provides an in-depth discussion of challenges and limitations, paving the way for future advancements. To systematically compare model outcomes, we also propose a dedicated evaluation framework. This framework enables an extensive and objective assessment of synthesized outputs, prioritizing two key properties: \texttt{quality} and \texttt{utility}. Quality is an intrinsic attribute independent of downstream tasks, encompassing aspects such as similarity, coverage, and diversity. It is assessed at both sample and distribution levels, considering various aspects such as statistical properties and temporal coherence. Utility refers to the capacity of synthetic data to serve as a replacement for or an augmentation of real training data in downstream predictive tasks.
\\
The key contributions and novelty of this work can be summarized as follows:
\begin{itemize}
\item We systematically assess the most relevant GAN and diffusion models, evaluating their capability to generate realistic and useful wearable sensor data.
\item We explicitly target complex sensor data generation, by combining multimodal inputs, LRD, and supervision signals (i.e., conditioning). To the best of our knowledge, this is the first study to offer a comprehensive and fair comparison of SoTA models in such a challenging benchmark. While crucial for wearable sensor data, our analysis also offers broader insights for optimizing and advancing TS generation methodologies.
\item We present a comprehensive, objective, and modality-agnostic evaluation framework, integrating both intrinsic quality assessment and downstream utility evaluation to ensure a holistic analysis of synthetic data performance.
\item We highlight the key challenges and limitations of existing approaches, offering guidelines and perspectives to drive the development of next-generation TS generative models.
\end{itemize}

The remainder of this paper is organized as follows. Section \ref{sec:rw} provides a review of generative models for TS synthesis, with a particular emphasis on GAN and diffusion models as leading approaches. It also examines challenges related to multi-modality, LRD, and conditional generation in the context of wearable sensor data, along with the open issue of synthetic TS assessment. Section \ref{sec:experiments} details the dataset selection and preprocessing, model training and inference procedures, and the evaluation framework for synthetic data. Section \ref{sec:result_discussion} provides a throughout discussion of the obtained results, while also addressing specific observations and potential limitations. Section \ref{sec:future_works} summarizes the key findings of our work and outlines future directions towards novel architectures for wearable sensor data generation.

\section{Motivations and Related Work}
\label{sec:rw}
The current landscape of generative models includes Variational Autoencoders (VAE) \cite{timevae}, Energy-Based Models (EBM) \cite{ebm}, Normalizing Flows (NF) \cite{normalizingflows}, GAN, and diffusion models.
Notably, Denoising Diffusion Probabilistic Models (DDPM) \cite{ho2020denoising} represent the leading diffusion paradigm, modeling both forward and reverse processes as Markov chains.
GAN and EBM do not explicitly model data distributions (i.e., implicit density models); instead, they transform samples drawn from a prior into realistic outputs, using the true data distribution to refine their estimates. In contrast, VAE, NF, and DDPM perform explicit modeling by mapping real data into a prior distribution (typically Gaussian) through an encoding process, then learn to decode samples drawn from this prior for data generation.
Although GAN and DDPM are not free from inherent drawbacks, they circumvent key issues that hinder the development and applicability of the other approaches. For example, in EBM, Markov Chain Monte Carlo (MCMC) methods are necessary to sample data from an approximation of the energy function - which is intractable for most models in its original analytical form - introducing a significant computational burden. On the other hand, NF require learning a sequence of invertible transformations to map noise to real data, necessitating efficient computation of the Jacobian determinant. This imposes restrictive network constraints and adds significant computational complexity. 
\\
The following subsections describe the specific design of GAN and DDPM for TS generation, highlighting their transition from the CV domain. For their fundamental principles, we refer the reader to the original works.
Then, we address the primary challenges associated with TS generation, particularly in the context of wearable sensor data, where these models often achieve satisfactory results only under strict constraints, thus limiting real-world applicability. Finally, we discuss the ongoing challenges of the definition of a consensus evaluation framework for synthetic TS data. 

\subsection{Overview of TS generation with GAN and Diffusion models}
\label{rw_1}
Since their introduction, GAN have largely dominated the CV domain. More recently, the emergence of diffusion models have set a new standard for perceptual quality \cite{rombach2022high}, while avoiding the optimization challenges inherent to adversarial training.
Motivated by their success in image generation, researchers have explored extending these generative frameworks to TS data. However, most efforts are direct adaptations from the CV domain.
Preliminary applications predominantly employ $2$-D Convolutional Neural Networks (CNN), with the U-Net architecture \cite{unet} serving as the standard backbone for the denoising network in DDPM. This often requires intermediate image-like representations of TS data, such as spectrograms \cite{kumar2019melgan,kong2021diffwave}. 
Although these transformations are, in some cases, easily invertible and capable of retaining most essential information (e.g., audio) while reducing the dimensionality of raw, high-rate waveforms, they inevitably introduce approximations of the original data. This may compromise the preservation of temporal dynamics and degrade the quality of the synthetic output.
\\
To address this limitation, further GAN exploited different network architectures, more suitable for sequential data modeling, such as Recurrent Neural Networks (RNN), particularly Long Short-Term Memory (LSTM) \cite{yang2023ts}, temporal CNN \cite{emotionalgan}, as well as hybrid architectures \cite{zhu2019electrocardiogram}.
On the other hand, the application of diffusion models to TS is still in a preliminary stage, mainly due to their relatively recent introduction. A recent survey \cite{yang2024survey} provides a detailed review of the literature in this area and indicates that most approaches rely on adaptations of the U-Net architecture for TS data, with only minor modifications and customizations, typically tailored to specific applications or underlying signals \cite{xiong2024patchemg, aristimunha2023synthetic}.
In contrast, the exploration of alternative architectures, such as Transformers, remains relatively limited, presenting promising opportunities for future research and advancements in this area.

\subsubsection{Multimodal TS generation}
TS generation, especially referring to physiological and behavioral sensor data, typically focuses on univariate signals, such as audio \cite{donahueadversarial}, photoplethysmogram (PPG) \cite{kiyasseh2020plethaugment}, and single-lead electrocardiogram (ECG) \cite{zhu2019electrocardiogram}.
However, most applications involves multiple sensing modalities.
In this context, research has primarily focused on task-agnostic generation, such as imputation \cite{zhou2024mtsci}, forecasting \cite{gao2023adversarial}, and denoising \cite{li2023descod}, which typically involve short-term sampling using extensive contextual information from observed (e.g., historical) data.
Conversely, multimodal generation is considerably more challenging due to complex correlations among and within heterogeneous TS, requiring accurate learning of their joint distributions from scratch.
Preliminary efforts have largely focused on MTS from the same modality, such as multi-lead ECG \cite{thambawita2021deepfake}, multi-channel EEG \cite{sharma2023medic}, and $3$-axis accelerometer data \cite{wang2022wearable}, which typically show strong correlations.
\\
Most solutions handle multimodal sensor data straightforwardly by concatenating different signals as separate channels within a unified input, which is processed as a whole. For instance, RCGAN \cite{esteban2017real} is an early approach to synthesizing multimodal Intensive Care Unit (ICU) data (heart rate, respiration rate, oxygen saturation, and blood pressure). However, it generates hourly summary statistics rather than high-frequency waveforms, which are subsequently evaluated using a downstream forecasting task to predict whether each signal will exceed predefined thresholds within the next hour.
Similarly, \cite{ehrhart2022conditional} proposed a hybrid LSTM-CNN GAN for the joint generation of stress-related electrodermal activity (EDA) and skin temperature. In this case, the task is considerably simplified, as the generated signals are limited to short-term stress responses (only $64$ data points) after external stimulation. Another example are GAN-based data augmentation (DA) approaches for HAR \cite{kang2022augmented}, which integrate Inertial Measurement Unit (IMU) data from accelerometers, gyroscopes, and magnetometers.
Similarly, DDPM typically treat multiple sensor data as separate input channels for the denoising network. However, this approach may not fully capture the intrinsic characteristics of multimodal distributions, limiting both cross-modal and intra-modal temporal coherence. As a result, the question remains open as to whether leading generative models can efficiently synthesize multimodal sensing data in parallel or whether new specialized approaches are necessary.

\subsubsection{Modeling LRD}
In the context of TS generation, capturing LRD remains a significant challenge for conventional Deep Learning (DL) models. CNN are inherently biased towards locality because of the limitations of their receptive fields. This presents a challenge when modeling a large context, since it necessitates a proportional increase in learnable parameters relative to the sequence length. In contrast, RNN are \emph{stateful} as they summarize the entire input into their hidden state, leading to slow training and vanishing gradient issues. Although specialized variants, such as dilated convolutions \cite{donahueadversarial} and SampleRNN \cite{mehri2017samplernn}, have been developed to mitigate these problems, their effectiveness is generally confined to limited contexts.
Recently, Transformers have become popular for sequential data processing due to their multi-head, self-attention mechanism \cite{vaswani2017attention}, which allows pairwise comparisons across all samples in a sequence. However, their inherent quadratic computational complexity restricts scalability.
To overcome this limitation, several efficient variants—often referred to as \texttt{xFormers}—have been proposed to substantially reduce the quadratic dependency on sequence length \cite{choromanski2021rethinking}.
\\
Eventually, Structured State Space Models (SSSM) \cite{gu2022efficiently} have been recently introduced as a new class of deep neural networks designed to effectively capture LRD. They can be considered as specific instantiations of both CNN and RNN, inheriting their efficiency during training and inference while addressing their main limitations. Specifically, SSSM offer a linear implementation for mapping inputs to outputs through hidden states, thus avoiding common optimization issues of classical RNN. As CNN, they represent a special case with an unbounded kernel, thus overcoming the limitations imposed by fixed receptive fields. SSSM have been applied both for autoregressive (AR) generation \cite{goel2022sashimi} and as \quotes{plug-and-play} backbones within non-AR frameworks. To the best of our knowledge, \cite{alcaraz2023sssd_ecg} is among the few studies that integrate SSSM as the denoising network within a DDPM, aiming to extend ECG generation up to $1$K samples.
As a result, accurate long-range generation of sensing data remains an open research challenge with substantial implications, as it can facilitate meaningful inference across a broad spectrum of downstream tasks.

\subsubsection{Conditional generation}
Generative models can be broadly classified as unconditional or conditional. Unconditional models generate output solely based on the learned distribution of the source data, without leveraging any external context. However, to avoid generating irrelevant samples, the source data must originate from a consistent distribution. In practice, this requires pre-selecting category-specific data prior to training, which reduces the dataset size and necessitates training separate model instances for each category.
In contrast, conditional models integrate additional context information to enable a more precise and fine-grained control over the statistical distributions of the generated data. Within the context of conditional GAN (CGAN), several strategies are employed to incorporate semantic information, yet no single approach has emerged as dominant. These strategies typically involve either concatenating conditioning information with the input of both generator and discriminator networks, or using an embedding layer to transform categorical labels into continuous vectors. Conditional batch normalization \cite{de2017modulating} is based on the adjustment of the normalization statistics of intermediate layers by using label information, allowing the network to modify its output based on the class labels. 
In contrast, the Auxiliary-Classifier GAN (AC-GAN) framework differs from CGAN by requiring the discriminator to assess both the realism of the sample and its class. Initially, this was achieved by integrating an additional classifier \cite{odena2017acgan}. However, more recent approaches have adopted a multi-task learning (MTL) strategy, enabling the discriminator to make both predictions using two classification heads \cite{li2022ttscgan}.
\\
In conditional diffusion models, label embeddings are typically combined with timestep embeddings, which indicate the current stage of the diffusion process, and then input to the denoising network. This allows the network to generate outputs that are both contextually relevant and temporally coherent. Two primary techniques for controlling inference in conditional DDPM are classifier guidance and classifier-free guidance \cite{ho2021classifierfree}. Classifier guidance employs a separate classifier, similar to AC-GAN, to predict class labels from intermediate noisy data. During inference, the classifier’s gradients are scaled and injected into the diffusion process to steer generation toward a target class. Although this provides precise control, it requires an additional model, increasing computational overhead and restricting control to classes seen during classifier training.
Classifier-free guidance, by contrast, integrates the guidance directly into the model. The model is trained on both conditional and unconditional noisy samples, and during inference, it combines their scores through scale factor to balance control precision and sample diversity. Due to its simplicity and efficiency, classifier-free guidance is now standard in SoTA conditional diffusion models, such as Stable Diffusion.
\\
Currently, the most used supervision signals are class labels, even if any other metadata might be incorporated. For instance, subject demographics as well as clinical information may be used to generate cohort-specific data. However, research in multi-label settings is currently limited. Few recent works, such as \cite{alcaraz2023sssd_ecg} and \cite{chung2023autotte}, addressed ECG generation conditioned on multiple statements, thereby expanding the range of unique label combinations. To this aim, they simply incorporated multi-label patient embeddings into their network architecture. However, high-dimensional (or even continuous) condition spaces introduce significant complexity that must be managed, as models need to learn more intricate input-output dependencies, which often require more data and computational resources. Additionally, as the observed conditions become sparser, more data gaps arise, which can potentially cause the model to perform poorly for unseen or infrequent conditions \cite{zhang2022mind}.
Therefore, advancing high-dimensional conditional data generation is crucial, particularly in critical application domains such as healthcare and precision medicine. This advancement can be exploited to reproduce and/or augment community- and individual-level datasets, paving the way towards personalized synthetic data generation to support more tailored solutions.

\subsection{Synthetic TS data evaluation}
\label{sec:TS_eval}
Currently, there is a broad consensus within the CV community on the evaluation of synthetic images.
Since realism and perceptual quality are relatively straightforward to assess in images, qualitative approaches rely on human annotations to evaluate these properties. Therefore, large-scale, cost-effective human-centered assessment has become a common practice, also facilitated by crowdsourcing platforms such as Amazon Mechanical Turk \cite{crowston2012amazon}.
In contrast, quantitative methods compare the statistical properties between synthetic and real images, with the Fréchet Inception Distance (FID) \cite{heusel2017gans} commonly used as the standard benchmark metric.
FID evaluates the similarity between the distributions of latent embeddings extracted from the deepest layer of a pre-trained Inception v$3$ network. This layer, close to the output, captures high-level image features (e.g., objects, shape, textures), allowing FID to quantify how closely real and synthetic data share these features.
\\
Unfortunately, adopting similar evaluations for TS data presents several challenges. Unlike images, TS data cannot be assessed from a psycho-perceptual standpoint by general users. In some cases, domain experts are required to evaluate waveform quality, such as cardiologists for artificial ECG data. In other cases, even specialized users may struggle to interpret the underlying content of TS data, for example, activity traces from IMU sensors. Consequently, human-centered evaluation in the TS domain faces considerable scalability barriers. Visual inspection is often used as a preliminary assessment of data generation quality; however, it is inherently subjective and time-consuming, limiting its applicability to a small fraction of the generated data. On the other hand, $2$-D distribution visualization using dimensionality reduction techniques are often used to provide an immediate qualitative indication of synthetic data representativeness.
For what concerns objective quality assessment, there is currently no FID-like authoritative benchmark for synthetic TS. Research in this area is still in its infancy, and a comprehensive analysis is needed to establish universal solutions to benchmark generative models across various types and tasks of TS.
\\
Moreover, evaluating TS is inherently multidimensional.
Stenger et al. \cite{stenger2024evaluation} recently proposed a taxonomy for synthetic TS assessment, based on the following properties: \texttt{fidelity}, \texttt{coverage}, \texttt{distribution matching}, \texttt{diversity}, \texttt{utility}, \texttt{novelty}, \texttt{privacy}, and \texttt{efficiency}. While an ideal metric would account for all criteria, this is impractical.
The first four properties represent different facets that collectively quantify the similarity between real and synthetic data, ultimately reflecting realism. In essence, synthetic data should resemble the patterns and statistical properties of real data while ensuring homogeneous coverage of the source data distribution, avoiding concentration in limited regions (i.e., low diversity or mode collapse). Utility is particularly important for synthetic data, and especially in human sensing applications, where their primary goal is to replace or augment real samples in low-data regimes to enable accurate predictive AI tasks.
\\
Moreover, wearable sensor data are often highly sensitive, and privacy breaches can have serious consequences. Although such data rarely contain directly identifiable information, they may encode unique biometric patterns capable of linking synthetic traces back to individual real samples. Consequently, it is critical for generative models to produce genuinely new instances rather than simply memorizing training data. Existing privacy protection strategies include differential privacy \cite{esteban2017real}, latent-space regularization or manipulation \cite{pennisi2023privacy}, and federated learning \cite{wijesinghe2024ps}. While privacy preservation and its analysis constitute a separate research line that goes beyond the scope of this work, the closely related concept of novelty remains essential for effective DA. However, current evaluation approaches rely on standard distance metrics between real and synthetic datasets, akin to similarity assessment yet with an opposing objective.
Lastly, efficiency refers to the computational cost of generating data. Although inference time is a major limitation for generative models, it is generally a secondary concern unless targeting edge-device applications.
\\
As a consequence, the multidimensional nature of the assessment landscape has produced a proliferation of disparate metrics. \cite{stenger2024evaluation} identified $83$ metrics across $56$ publications, revealing a fragmented ecosystem in which each study adopts its own subset of evaluation criteria. Notably, most metrics are rarely reused, and only a small fraction are consistently applied. Although some were introduced recently, these findings suggest that many metrics have had little to no impact on the research community and point to a \textit{\quotes{do as you like}} approach that hinders progress toward a unified evaluation standard. Furthermore, the diversity of sensing modalities in wearable systems also introduces data-specific assessment. For example, synthetic ECG may be evaluated via heart rate (HR) and heart rate variability (HRV), while EDA is evaluated through tonic and phasic statistics.
\\ 
Given these challenges and limitations, establishing a data-agnostic, comprehensive, and objective assessment procedure is essential to improve the comparison of TS generative models. Therefore, building on key insights from \cite{stenger2024evaluation}, we introduce a synthetic TS evaluation framework specifically designed for wearable sensor data as input for human sensing applications.
This framework integrates the most relevant metrics from the literature, allowing both intrinsic, task-independent quality assessment and utility evaluation in downstream predictive tasks that reflect the intended use of the real data.

\section{Methods}
\label{methods} 
This study evaluates the leading SoTA GAN as well as some latest diffusion models for TS data generation, focusing on recent, well-established approaches. Selection is guided by two key criteria: (i) specific design for TS generation suitable for mobile and wearable sensor data, (ii) publicly available implementations to ensure experiment reproducibility and fair comparison. Through these models, we aim to deliver a comprehensive analysis highlighting the limitations and challenges of generating synthetic datasets derived from real-world studies.
In the following subsections, we present the selected GAN and diffusion models, outlining their design, architecture, and relevance to our task. We then describe the design and rationale of the proposed evaluation framework, aimed at assessing the strengths and limitations of these generative models.
 
\subsection{Selected TS GAN}
\label{sec:ts-gan}
As noted by Brophy et al. \cite{brophy2023generative}, there is a limited availability of high-quality GAN specifically designed for TS data generation.
Among these, \texttt{TimeGAN}\footnote{\url{https://github.com/jsyoon0823/TimeGAN}} \cite{yoon2019time} is the first model specifically designed to preserve temporal dynamics, by combining the flexibility of unsupervised learning offered by GAN with the control of supervised training in AR models.
The model consists of two main components: an autoencoder (AE) and a standard GAN architecture consisting of a generator ($G$) and a discriminator ($D$). These components are \emph{jointly} trained such that TimeGAN simultaneously learns to encode data in a lower dimensional space, generate latent representations, and synchronize the stepwise dynamics of both real and synthetic embeddings to create similar temporal transitions. As a result, the overall training procedure involves the optimization of a weighted combination of the following loss functions: $1$) a reconstruction loss to ensure an accurate and reversible mapping between original data and their latent vectors, a $2$) a standard adversarial loss to encourage realism of synthetic embeddings, and $3$) a supervised AR loss that has a constraining effect on the sample-wise dynamics of the generator.
\\
However, modeling TS data requires learning patterns across different timescales, including both short- and long-term dependencies. In this context, \texttt{WaveGAN} \cite{donahueadversarial} has been introduced for unconditional audio generation and is based on DCGAN \cite{radford2015unsupervised}, a popular GAN framework
for image synthesis. In DCGAN, the $G$ employs transposed convolutions (or deconvolutions) to iteratively upsample feature maps, producing outputs with the same (or even higher) resolution as the input images. WaveGAN adapts this architecture to $1$-D data by flattening the network and replacing transposed convolutions with dilated convolutions \cite{oord2016wavenet}, exponentially increasing the receptive field with only a linear increase in depth.
An enhanced variant, \texttt{WaveGAN$^*$} \cite{thambawita2021deepfake}, supports multi-channel inputs and features a deeper architecture with additional deconvolution blocks in both $G$ and $D$ to capture more complex signal features.
\\
\texttt{Pulse2Pulse} (\texttt{P2P}) \cite{thambawita2021deepfake} is another prominent GAN framework designed for TS data, with specific application to multi-lead ECG. This framework introduces for the first time the U-Net \cite{ronneberger2015u} as the $G$ network, adapting it to TS data through the use of $1$-D convolutional filters.
On the other hand, P$2$P shares the same $D$ architecture of WaveGAN$^*$. 
More recently, conditional variants of P$2$P and WaveGAN$^*$, referred to as \texttt{P2P$_{COND}$} and \texttt{WaveGAN$^*$$_{COND}$}, have been introduced in \cite{alcaraz2023sssd_ecg}. These models incorporate conditional batch normalization into each convolutional layer, allowing the network’s internal scaling and shift parameters to be influenced by external labels. Label vectors are firstly transformed into continuous embeddings, which are then added to the convolutional outputs.
\\
Finally, \texttt{TTS-GAN} \cite{li2022ttsgan} is a leading framework for TS generation built on a pure Transformer-encoder architecture. Inspired by Vision Transformers (ViTs) \cite{dosovitskiy2021an}, it adapts to TS data by representing each input as a $C \times H \times W$ tuple, where $C$ denotes the number of channels, $H$ represents the height (equal to $1$ for TS data), and $W$ corresponds to the sequence length. The model segments each sample into non-overlapping, fixed-length patches (a process known as \emph{patchification}), then applies positional encoding to each patch. The same authors also introduced a conditional variant later on, \texttt{TTS-CGAN}\footnote{\url{https://github.com/imics-lab/tts-cgan}} \cite{li2022ttscgan}, after experimenting with different embedding strategies. Their best-performing conditioning approach involves concatenating the label embedding with the generator input and adding a further classification head to the discriminator, thereby incorporating a categorical cross-entropy term within the discriminator objective.

\subsection{Selected TS DDPM}
\label{sec:ts-ddpm}
In recent years, several studies have investigated TS generation using DDPM \cite{yang2024survey}. However, a closer examination of the literature reveals that most approaches adopt the U-Net architecture with little or no modification, applying it to different signals. As a result, the choice of the denoising network remains largely inherited from the CV domain.
Among these, \texttt{Biodiffusion}\footnote{\url{https://github.com/imics-lab/biodiffusion}} \cite{li2024biodiffusion} stands out as one of the most recent publicly available models, specifically designed for biomedical TS generation. Therefore, it can serve as a strong baseline in our reference domain.
The adaptation of \texttt{Biodiffusion} to TS data is achieved using flattened convolutions, augmented by multi-head attention layers at the end of each residual block to help the model capture salient features and complex dependencies, improving reconstruction.
In \cite{alcaraz2023sssd_ecg}, the authors presented \texttt{Structured State Space Diffusion} for multi-lead ECG generation (SSSD-ECG)\footnote{\url{https://github.com/AI4HealthUOL/SSSD-ECG}}.
The key innovation lies in the integration of SSSM as the denoising backbone of DDPM, enabling more effective capture of LRD.
Please note that both WaveGAN$^*$$_{COND}$ and P$2$P$_{COND}$ have been developed as baselines and are available in the same code repository.
\\
Eventually, Peebles et al. \cite{peebles2023scalable} have recently introduced a novel class of diffusion models called \texttt{Diffusion Transformers} (\texttt{DiT}), which leverage Transformer architectures. In their approach, they trained a DDPM on low-dimensional image embeddings obtained through a pre-trained VAE, replacing the traditional U-Net backbone with a deep stack of Transformer blocks that processing latent patches. This method sets a new benchmark for high-resolution image synthesis, while also demonstrating favorable scaling properties in terms of model complexity vs. sample quality. Building on these insights, and considering the inherently sequential nature of TS data, few and very recent studies extended the application of DiT to this domain \cite{yuan2024diffusionts,sikder2025transfusion}.
Although preliminary results on simple datasets are promising, the adaptation of \texttt{DiT} to TS remains at an early stage. Key challenges include selecting suitable models for general, low-dimensional TS representation learning and determining optimal patch lengths and attention modules, with the aim of improving sample quality while managing the high computational cost of Transformers, particularly for long and multivariate sequences. Consequently, we selected BioDiffusion and SSSD as our reference diffusion models, leaving the exploration of DiT for future work.

\subsection{Evaluation framework}
\label{sec:eval_framework}
Based on the challenges discussed in Section \ref{sec:TS_eval}, we propose a comprehensive, modality-agnostic evaluation framework that prioritizes both the similarity and utility of synthetic data, with the ultimate goal of enabling fair performance comparisons across different TS generative models. For this purpose, our framework defines two specific evaluation procedures, namely task-dependent and task-independent, which are described in the following subsections. In both cases, we selected the most relevant metrics from the current literature, as highlighted in \cite{stenger2024evaluation}.

\subsubsection{Task-independent evaluation}
When handling MTS, many approaches commonly perform a global evaluation that aggregates all channels. However, this can obscure signal-specific failures or suboptimal performance. To address this, we chose a modality-specific analysis to provide a fine-grained assessment of data quality and identify potential disparities. We also performed an intra-class evaluation; however, since inter-class differences may naturally arise from variations in data sizes and underlying pattern complexities, we report class averages in our results.
\\
First, we performed a preliminary qualitative assessment visualizing data distributions in a $2$-D space using t-distributed Stochastic Neighbor Embedding (t-SNE) as a dimensionality reduction technique. This visualization highlights the degree of overlap between real and synthetic data, providing an immediate insight into disjoint distributions, mode collapse, and other failure cases.
We then computed a set of distance metrics to evaluate the similarity between individual synthetic and real sequences, including cosine distance (CD), correlation distance (CrD), and Euclidean distance (L$2$). Although these metrics operate at sample level, we averaged across all possible real–synthetic pairs to obtain a robust global measure at the distribution level.
Both cosine and correlation distances fall within the range $[0,2]$, where $0$ indicates perfect similarity (correlation), $1$ indicates orthogonality (no correlation), and $2$ indicates perfect dissimilarity (anti-correlation).
However, applying such distance metrics to long sequences often suffers from the \textit{\quotes{curse of dimensionality}}, leading to less meaningful comparisons. As the dimensionality increases, pairwise distances become more similar, reducing their discriminative power and thus hindering their direct applicability to long, raw TS data. Therefore, following \cite{li2022ttsgan}, we extracted a set of statistical features for each sequence and computed the distance between the resulting low-dimensional vectors.
\\
We also incorporated additional measures to complement statistical similarity. Specifically, we calculated the average pairwise Dynamic Time Warping Distance (DTWD) \cite{berndt1994dtw} to assess temporal coherence. DTW similarity is determined by identifying the optimal alignment between two sequences, represented by a warping path through the distance matrix that minimizes the total alignment cost. This minimum total cost is usually referred to as the DTWD.
Furthermore, we used the Maximum Mean Discrepancy (MMD) \cite{gretton2012kernel}, a kernel-based statistical test used to determine whether two sets of samples are drawn from the same distribution. MMD operates at the distribution level, utilizing a kernel function $K$ (an exponentiated quadratic kernel in our case) to quantify the similarity between real and synthetic datasets.  We also normalized the kernel function so that its output ranges between $0$ and $1$, thus resulting in MMD values between $0$ and $2$ ($0=$ perfect similarity, $1=50\%$ similarity, $2=$ perfect dissimilarity).
Finally, we computed the discriminative score (DS), defined as the accuracy of a post-hoc classifier trained to distinguish real from synthetic data, with the ideal value corresponding to random guessing. In the absence of benchmark models specifically designed for this task, previous studies typically employed a single, simple network—often a shallow GRU-based RNN—for score computation \cite{yoon2019time,li2022ttsgan,coscigan2022,psa-gan}. In contrast, our analysis uses a more diverse set of DL classifiers with varying complexity, namely Multi-Layer Perceptron (MLP), Autoencoder, CNN, Fully Convolutional Networks (FCN), hybrid ConvLSTM, and ResNet.
Given a real dataset and its synthetic copy, we first split the two datasets into training-validation-test partitions with a $70$-$10$-$20\%$ proportion using stratification to preserve class ratio(s), and concatenated the corresponding real and synthetic sets. Each model was trained with a fixed architecture (no hyperparameter tuning) for $100$ epochs with a batch size of $32$, using Adam optimizer (learning rate $= 0.001$, $\beta _1 = 0.9$, $\beta _2 = 0.999$). Finally, we selected the model checkpoint corresponding to the epoch with the lowest binary cross-entropy loss and used it to evaluate the accuracy on the test set. The discriminative score is reported as $\|0.5-accuracy\|$, so it is bounded between $0$ and $0.5$. We averaged the scores across all classifiers to obtained an overall robust assessment.
\\
Beyond similarity assessment, we also evaluated the diversity of the generated data. Given the lack of standardized metrics for this purpose, we extended beyond conventional intra-class distance approaches \cite{norgaard2018synthetic} to measure synthetic-to-synthetic similarity by introducing spectral entropy (En) as a measure of distributional disorder, serving as a proxy for in-distribution diversity. Therefore, we computed entropy for both real and synthetic datasets separately and reported their absolute difference.
\\
As a result, our task-independent evaluation includes $7$ different metrics (plus visual assessment), thus offering extensive coverage. For all metrics, lower values are always the better.

\subsubsection{Task-dependent evaluation}
After evaluating the intrinsic quality of the synthetic data, we also evaluated their utility for downstream classification tasks, including both binary and multi-class settings, which mirror the original tasks of the corresponding source datasets.
Our primary objective is to determine the extent to which synthetic data can serve as a replacement for real training data, as well as their effectiveness when integrated into hybrid training sets for DA. In the first scenario, we seek for comparable performance while allowing for a limited performance drop, while we expect performance improvements in the second case.
Due to the absence of benchmark predictive models, we reused the same DL models previously introduced to compute the discriminative score, as outlined in the previous subsection. However, to explicitly assess the impact of multi-modality on predictive performance, all sensing modalities were jointly provided as inputs to the classifiers. Due to the class imbalance observed across all datasets (see Table \ref{tab:dataset-table}), we selected the Area Under the Receiver Operating Characteristics Curve (AUROC) as reference evaluation metric, using a weighted average for multi-class settings (CASE). Using the same training-validation-test proportion for both the real dataset and its synthetic counterpart described in the discriminative score computation, and we evaluated each model with the following combinations of real and synthetic samples:
\begin{itemize}
    \item Train on Real, Test on Real (TRTR): it serves as baseline for comparing the other approaches involving synthetic data;
    \item Train on Synthetic, Test on Real (TSTR): models are trained and validated exclusively on synthetic data, with testing on real data only.
    \item DA: real data are augmented with synthetic data to create hybrid training and validation sets. Specifically, we implemented three distinct policies that reflect practical usage of synthetic data in predictive AI:
    \begin{itemize} 
        \item \textit{Balance}: Synthetic instances are added only to the minority class(es) to balance the real training and validation sets. 
        \item \textit{Double}: The real and synthetic training and validation sets are simply concatenated, doubling their sizes while preserving original class distribution.
        \item \textit{Balance$+$Double}: 
        The real training and validation sets are first balanced, then their size is doubled by adding synthetic data to all classes. 
    \end{itemize}
\end{itemize}
This choice allows to evaluate how the progressive inclusion of synthetic data affects classification performance.
For both TSTR and all DA policies, we calculated the delta relative to the TRTR baseline for each model. The differences were then averaged among classifiers to obtain a global score. 

\begin{table}[ht]
\centering
\caption{Overview of the proposed evaluation metrics for synthetic wearable sensor data.}
\label{tab:eval-table}
\resizebox{\linewidth}{!}{
\begin{tabular}{|c|c|c|c|}
\hline
\textbf{Evaluation metrics}      & \textbf{Data Properties} & \textbf{Granularity} & \textbf{Task-dependent}\\ \hline
Distribution visualization & Similarity, Coverage, Diversity & Distribution & $\times$ \\ \hline
Avg. pairwise distances& Similarity& Sample& $\times$ \\ \hline
MMD, discriminative score& Similarity& Distribution & $\times$ \\ 
\hline
Spectral entropy & Diversity &Distribution & $\times$ \\ \hline
TSTR, DA & Utility & Distribution & $\checkmark$  \\ \hline
\end{tabular}
}
\end{table}

\section{Experiments}
\label{sec:experiments}
\subsection{Data selection and curation}
\label{sec:data_curation}
Multimodal wearable datasets are notably scarce in major public repositories (e.g., Physionet). This is due to ethical, privacy, and monitoring constraints, thus making generative modeling for multimodal signals particularly valuable. For this study, we selected the following reference datasets: \texttt{WESAD} \cite{wesad}, \texttt{SWELL} \cite{swell}, and \texttt{CASE} \cite{case}, which are widely used for benchmarking stress and emotion modeling and recognition tasks. These datasets include multimodal recordings from one or more wearable devices. While primarily designed for affective computing, the collected signals are also relevant to other domains, such as mental health, sleep research, and human-AI interaction, allowing our evaluation to be representative for broader applications.
We focused on physiological signals common to all datasets, namely ECG and EDA, which exhibit complementary characteristics. ECG shows quasi-periodic patterns with distinct waveforms, while EDA is a slowly varying signal with a tonic baseline and phasic peaks in response to external stimuli. Moreover, both signals serve as key biomarkers of the Autonomic Nervous System (ANS) response to stress and emotions \cite{sharma2012objective}, making them ideal reference signals for the selected  downstream tasks.
\\
In each dataset, signals are all synchronized yet recorded at different sampling rates. Specifically, in WESAD, chest ECG is sampled at $700$Hz, while wrist EDA is sampled at $4$ Hz. In contrast, in SWELL and CASE both signals are sampled at the same rate of $2048$Hz and $1000$Hz, respectively. To ensure consistency, we resampled all signals to a common rate of $100$Hz for WESAD and CASE, and $128$Hz for SWELL, using downsampling or upsampling. We did not scale SWELL data to $100$Hz to avoid introducing interpolation. This resampling rate is appropriate for ECG, as previous studies have shown that higher rates (e.g., $500$Hz) do not offer significant improvements in signal quality and classification performance \cite{mehari2022advancing}.
On the other hand, EDA dynamics is typically below $5$Hz \cite{society2012publication}, so noise artifacts may be added. To mitigate this issue, we applied a low-pass filter with a $5$Hz cutoff, obtaining a cleaned EDA version.
Next, we divided each signal according to the different phases of the corresponding monitoring protocol, then we further segmented each phase into non-overlapping $10$-s windows, resulting in sequences of $1000$ and $1280$ data points, respectively. This approach enables testing LRD modeling performance and aligns with previous studies targeting longer sequences \cite{alcaraz2023sssd_ecg}. 
\\
We labeled each dataset to enable conditional data generation and assess the effectiveness of synthetic data in downstream classification tasks. We used only class labels as conditioning information, excluding additional metadata (e.g., demographics) to limit the scope of conditioning and prevent overloading the models with additional complexity beyond the already challenging tasks of multimodal and long-range data generation. Moreover, defining a consistent set of metadata to obtain a uniform conditioning across datasets, beyond sex and age group, was not feasible. As a result, we implemented binary stress detection for WESAD and SWELL and multi-class valence-arousal detection for CASE. In WESAD, following the original study, data windows from baseline and amusement periods were labeled as \texttt{no stress}, while those from the Trier Social Stress Test were labeled as \texttt{stress}. We excluded recovery phases due to their intermediate stress nature, making them unsuitable for binary stress classification. For SWELL, neutral working periods were labeled as \texttt{no stress}, while periods involving time pressure and interruption stressors were labeled as \texttt{stress}.
In the CASE dataset, participants provided continuous self-assessments of valence and arousal using a joystick-based annotation interface. To classify the data, we averaged the scores within each segment and applied a threshold of $5$ (on a $0$–$10$ scale) to categorize them into four combinations of \texttt{Low}/\texttt{High} valence and arousal.
Table \ref{tab:dataset-table} provides an overview of the datasets following our pre-processing pipeline, detailing the number of classes, subjects, instances, and the ratio between the majority and minority class(es).

\begin{table}[ht]
\centering
\caption{Overview of the processed source datasets.}
\label{tab:dataset-table}
\resizebox{\linewidth}{!}{
\begin{tabular}{|c|c|c|c|c|c|}
\hline
\textbf{Dataset} & \textbf{Task} & \textbf{\# subjects}& \textbf{\# classes} & \textbf{\# samples} & \textbf{Class ratio(s)} \\ \hline
WESAD & stress detection & 15 & 2 & 3924  & 2.9         \\ \hline
SWELL & stress detection& 25& 2 & 17550 & 1.6          \\ \hline
CASE  & valence-arousal level detection & 30 & 4 & 7350  & 3.5/4.5/11.1 \\ \hline
\end{tabular}
}
\end{table}

\begin{table}[ht]
\centering
\caption{Hyperparameter search space.}
\label{tab:hyperparams-table}
\begin{tabular}{|c|c|}
\hline
\textbf{Hyperparameter} & \textbf{Values} \\ \hline
Batch size & [8, 16, 32] \\ \hline
Training epochs & [100, 300, 500] \\ \hline
Data normalization & [Y, N] \\ \hline
Network backbone (TimeGAN only)& [GRU, LSTM, Bi-LSTM]
\\ \hline
\end{tabular}
\end{table}

\subsection{Model configurations}
\label{subsec:config}
Considering the previously selected models, we made only minor modifications to their original implementations in order to accommodate input data dimensions. Specifically, we adjusted the input layer of each model to accept two fixed-length input channels (ECG and EDA). For P$2$P${COND}$ and WaveGAN$^*$${COND}$, we slightly modified the upsampling and stride factors of the internal deconvolution blocks to match our sequence lengths.
\\
We performed a grid search to tune some of the most impactful general hyperparameters, including batch size, number of training epochs, and data normalization, as summarized in Table \ref{tab:hyperparams-table}. This resulted in a total of $18$ trials per model. For TimeGAN, we also tested $3$-layer GRU, LSTM, and Bi-LSTM architectures for $G$, $D$, and AE components, following the original implementation, yielding $54$ total trials. Since TimeGAN is an unconditional model, it was used as a baseline for comparison with conditional approaches, and each configuration was trained separately to generate class-specific data. 
Unless stated otherwise, we kept all other model-specific hyperparameters at their original defaults, as provided in the corresponding code repositories. While tuning other general hyperparameters, such as learning rates, could offer minor improvements, this is beyond the scope of our study, as it is unlikely to significantly affect model performance. It is also worth noting that most hyperparameters follow common defaults from current literature.
For example, all GAN employ the Wasserstein distance with gradient penalty (WGAN-GP) objective to mitigate training instability and mode collapse, with the gradient penalty factor set to $\lambda_{GP}=10$ according to the original implementation \cite{gulrajani2017improved}.
Diffusion models use $T=1000$ timesteps, the de facto standard sufficient to fully convert each data instance into noise. At this scale, the choice of the noise scheduler has minimal impact, so the default is used; its effect becomes significant only when the number of denoising steps has to be reduced to enable faster sampling. The guidance scale factor, which controls the adherence of generated samples to the supervision signal, is set to $3.0$ based on prior empirical studies indicating that this value can offer a good trade-off between sample diversity and class fidelity.
\\
Model checkpoints were saved every $5$ epochs, and the best checkpoint was selected post-hoc according to distinct criteria. For GAN, selection was based on the minimum generator loss, whereas for DDPM, the checkpoint minimizing the distance (MAE or MSE, depending on the implementation) between the denoised predictions and ground truth data was chosen. Although this provides a straightforward and generalizable model selection strategy, we explicitly discuss its intrinsic limitations and associated challenges in Section \ref{sec:limitations}.
 Using the selected model checkpoints, we first generated a synthetic digital twin of the source datasets, with same size and class distribution. We used this synthetic dataset for task-independent evaluation, TSTR, and DA in the \textit{Double} mode. Subsequently, we conducted a second round of inference in order to assess DA in \textit{Balance} and \textit{Balance$+$Double} settings, which require additional synthetic samples for each class.
\\
Model training, inference, and synthetic data evaluation have been conducted on a single node equipped with an NVIDIA A$100$ $80$GB GPU.

\begin{table}[ht]
\centering
\caption{Overview of failure cases across different models and datasets. \textit{U-} = unnormalized training data; \textit{N-} = normalized training data; \textit{Bxx} = batch size. Number of training epochs is $500$ in all cases.}
\label{tab:invalid_configs}
\resizebox{\linewidth}{!}{
\begin{tabular}{|c|c|c|c|c|c|c|c|}
\hline
& & \multicolumn{6}{c|}{\textbf{Configuration ID}}\\
\hline
\textbf{Model}& \textbf{Dataset} & \textbf{U-B8} & \textbf{U-B16}& \textbf{U-B32}&\textbf{N-B8} & \textbf{N-B16}& \textbf{N-B32}\\ \hline 
TimeGAN & WESAD & $\times$ & $\times$&$\times$ &$\times$ & $\times$ & $\times$ \\ 
\hline
TimeGAN & SWELL & $\times$ & $\times$&$\times$ &$\times$ & $\times$ & $\times$ \\ 
\hline
TimeGAN & CASE & $\times$ & $\times$&$\times$ &$\times$ & $\times$ & $\times$ \\ 
\hline
P$2$P$_{COND}$ & WESAD & $\times$ & $\times$& $\times$ & \checkmark & \checkmark & \checkmark \\
\hline
P$2$P$_{COND}$ & SWELL & $\times$ & $\times$& $\times$ & $\times$ & $\times$ & $\times$ \\ 
\hline
P$2$P$_{COND}$ & CASE & $\times$ & $\times$& $\times$& \checkmark & \checkmark & \checkmark \\ 
\hline
WaveGAN$^*$$_{COND}$ & WESAD & $\times$ & $\times$& $\times$ & \checkmark & \checkmark & \checkmark\\ 
\hline
WaveGAN$^*$$_{COND}$ & SWELL & $\times$ & $\times$& $\times$ & \checkmark & \checkmark & \checkmark\\
\hline
WaveGAN$^*$$_{COND}$ & CASE & $\times$ & $\times$& $\times$& \checkmark & \checkmark & \checkmark 
\\ 
\hline
\end{tabular}
}
\end{table}

\section{Results and Discussion}
\label{sec:result_discussion}
We observed consistent improvements for $N>300$ training epochs; therefore, results are reported for $N=500$ hereafter.
As a first step, we performed a preliminary evaluation of each model configuration to assess its validity, summarizing the failure cases in Table \ref{tab:invalid_configs}. Such initial findings indicate that TimeGAN fails to generate meaningful results, producing extremely low-quality data for both signals and causing significant degradation in downstream task performance. This outcome can be explained by two main factors. First, as an unconditional generative model, TimeGAN lacks the ability to leverage transfer learning effects between classes. In contrast, conditional models often learn a shared latent space that may capture common information among classes, facilitating knowledge transfer and enhancing data synthesis across different categories. Moreover, unconditional models face greater challenges when modeling minority classes, as insufficient data can hinder the training of a robust model. This is consistent with previous studies demonstrating that conditional generative models, which use data labels, outperform their unconditional counterparts \cite{bao2022why}. Second, by focusing mainly on stepwise dynamics, TimeGAN fails to capture long-term temporal correlations and inter-channel dependencies, thus degrading the overall temporal coherence of the generated sequences.
\\
For P$2$P$_{COND}$, model configurations trained on raw, unnormalized data cannot generate class $0$ (\quotes{No Stress} for WESAD and SWELL, \quotes{Low Valence-Low Arousal} for CASE) of both ECG and EDA signals across all datasets, showing a pronounced mode collapse. For SWELL, configurations trained with normalized data also failed to produce meaningful ECG data for the same class.
Consequently, our analysis is limited to configurations using normalized data, with results reported only for class $1$ (\quotes{Stress}) ECG data in the SWELL dataset. We observed identical failure cases with WaveGAN$^*$$_{COND}$, highlighting a similar behavior between the two models.

\begin{figure*}[ht] 
    \subfigure[WESAD, ECG]{ \includegraphics[width=0.45\textwidth]{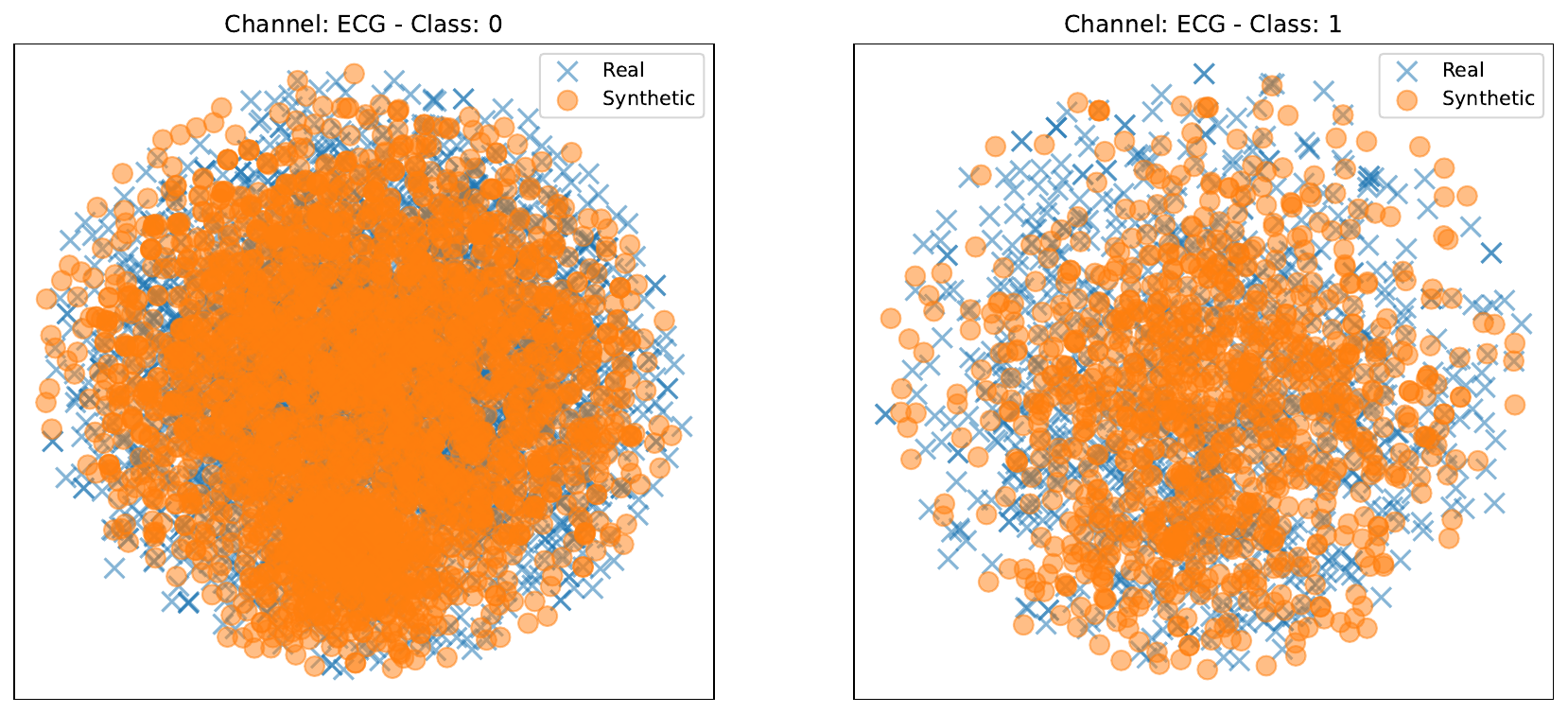}}
    \hfill
    \subfigure[WESAD, EDA]{\includegraphics[width=0.45\textwidth]{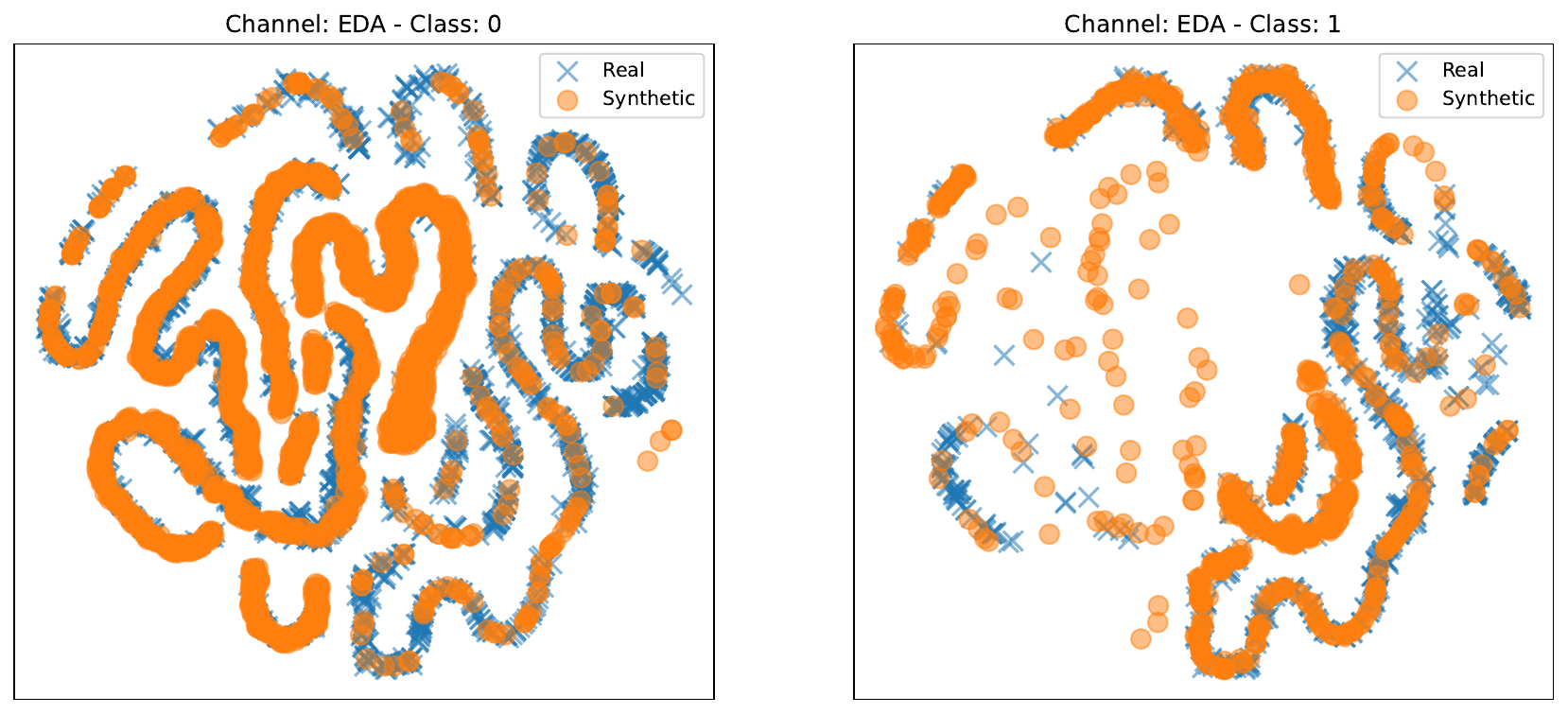}}
    \\
    \subfigure[SWELL, ECG]{ \includegraphics[width=0.45\textwidth]{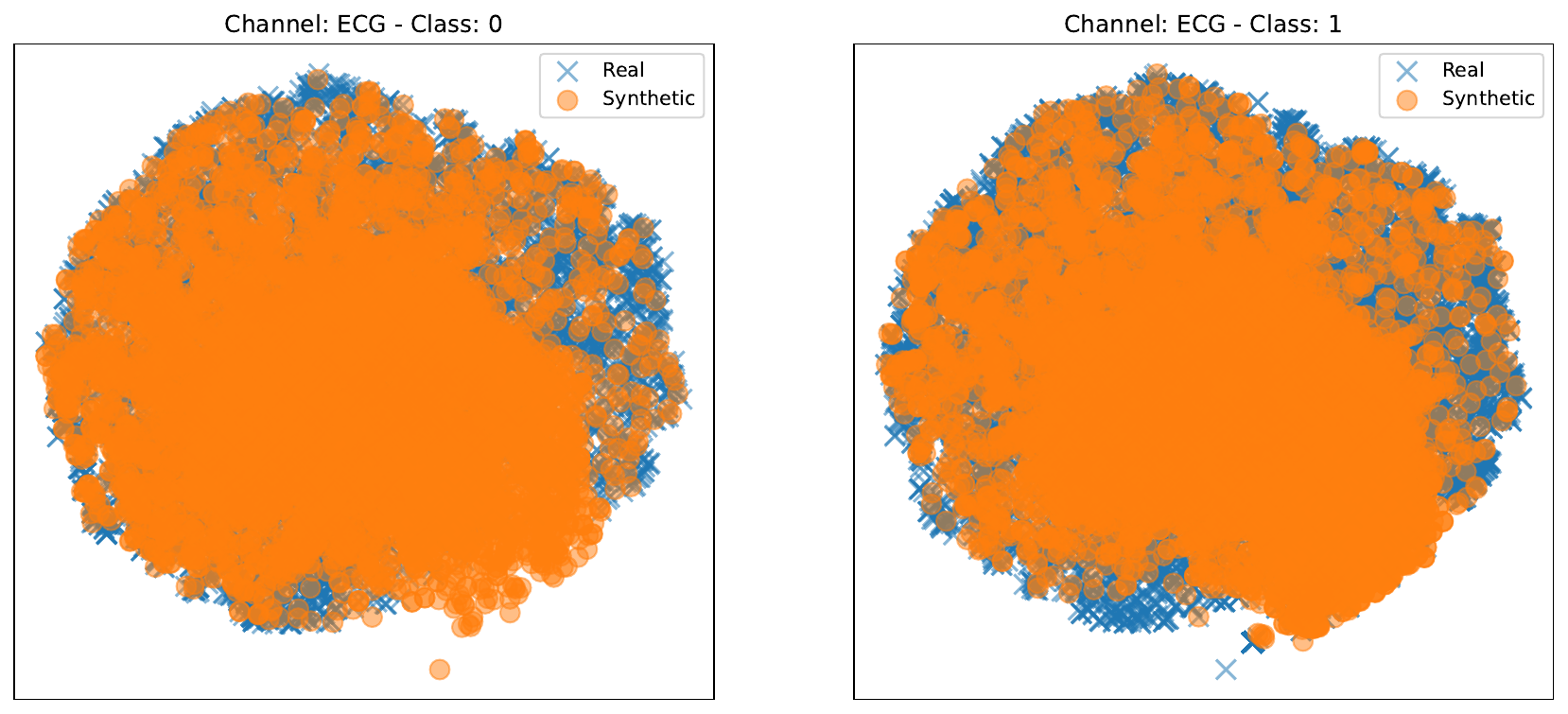}}
    \hfill
    \subfigure[SWELL, EDA]{\includegraphics[width=0.45\textwidth]{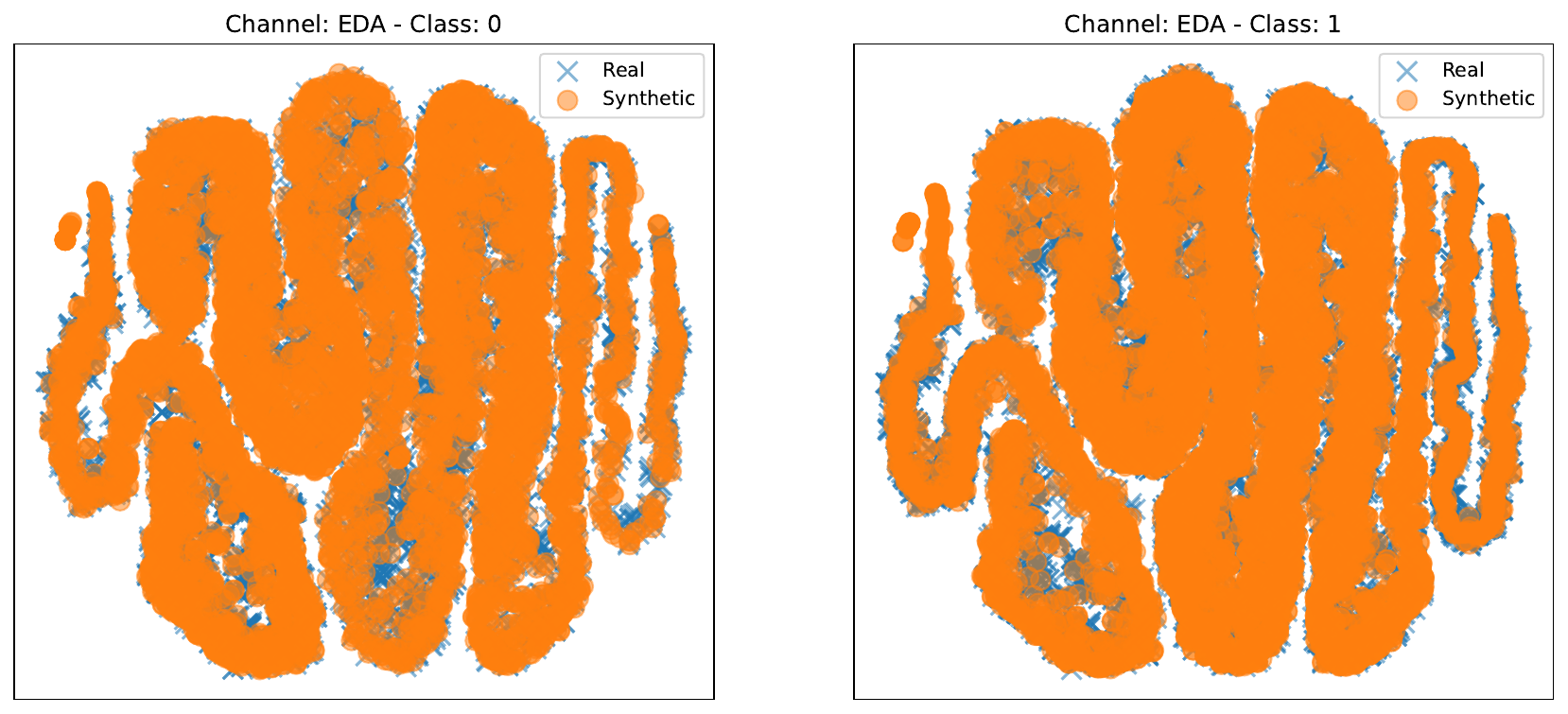}}
    \\
    \subfigure[CASE, ECG]{ \includegraphics[width=0.45\textwidth]{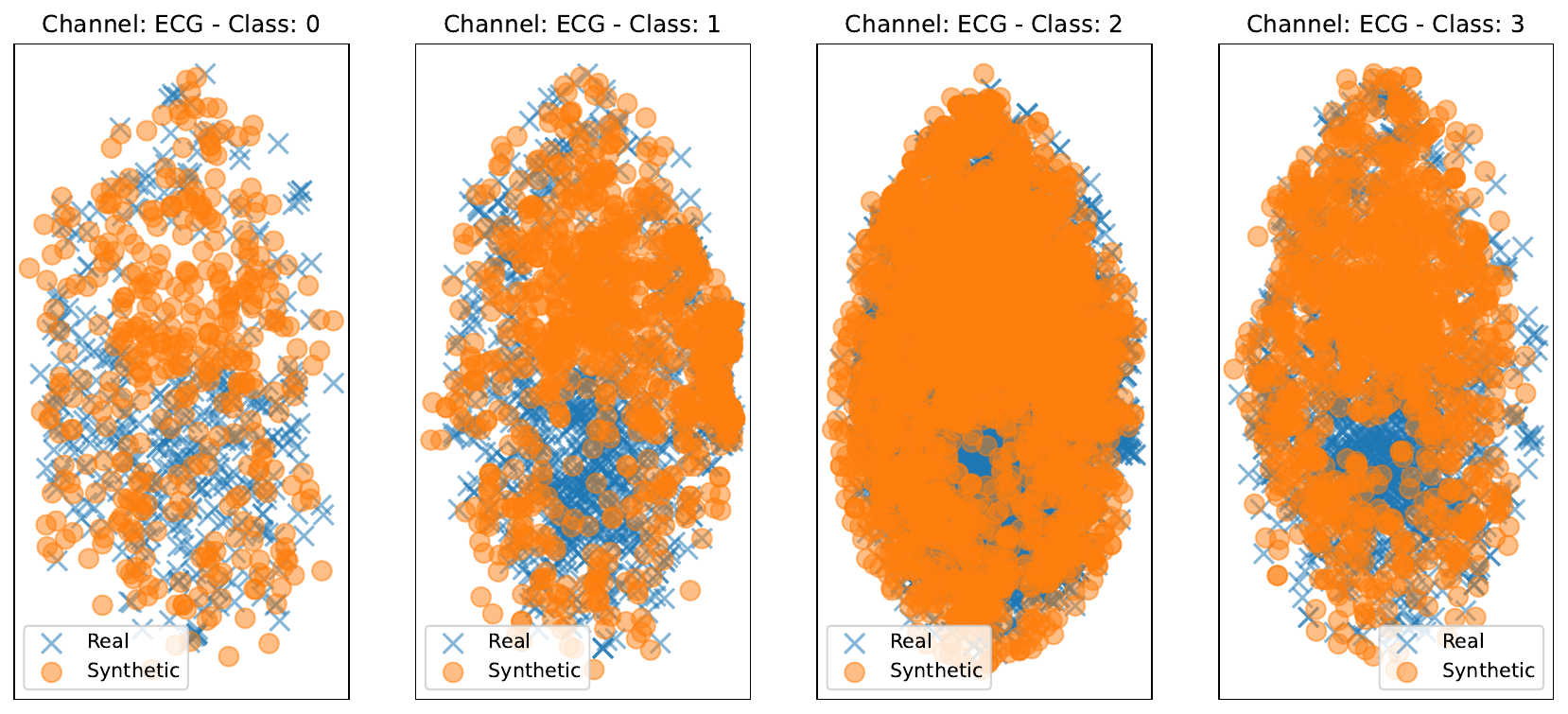}}
    \hfill
    \subfigure[CASE, EDA]{\includegraphics[width=0.45\textwidth]{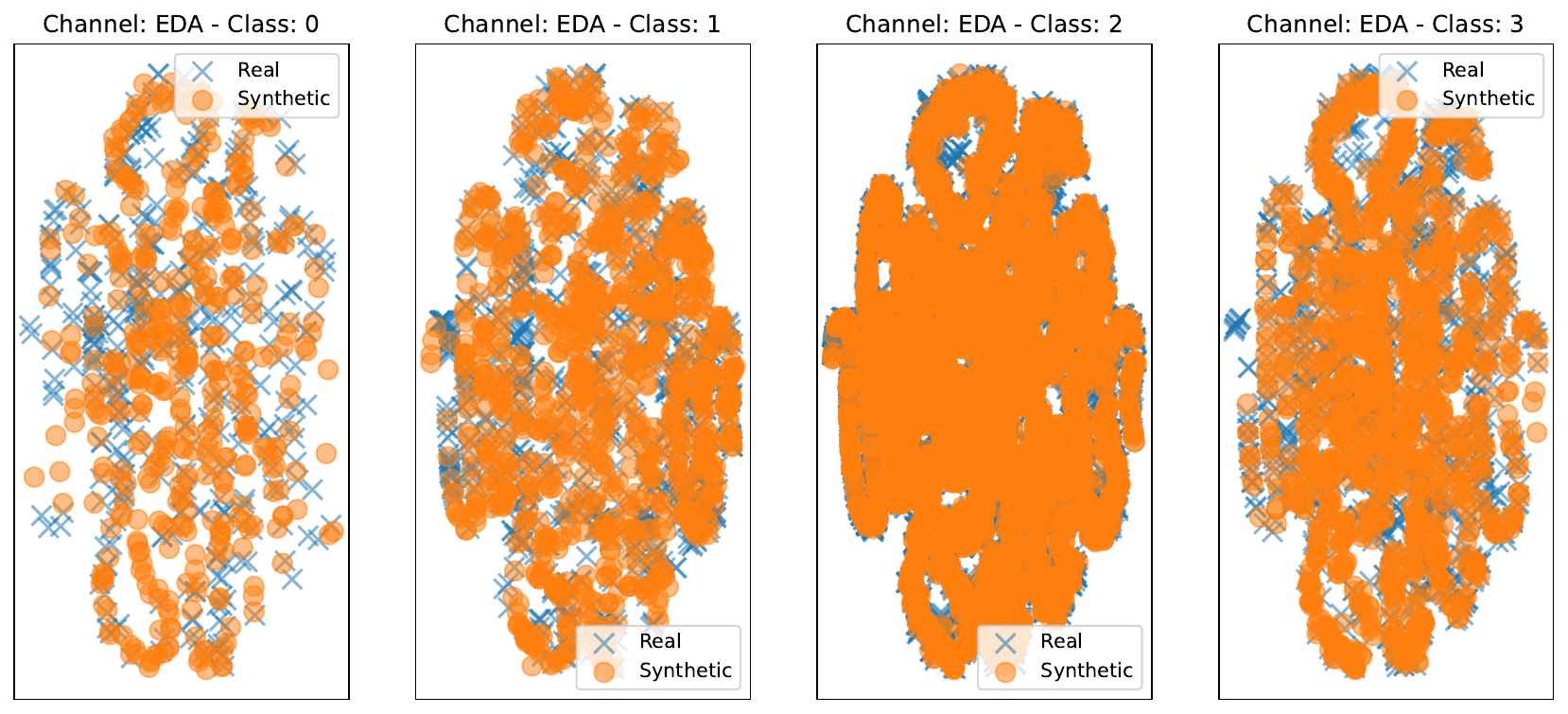}}
    \label{fig:tsne}
    \caption{Signal- and class-specific t-SNE visualizations of real vs. synthetic distributions for the top-performing BioDiffusion model for each dataset.}
\end{figure*}

\begin{table*}[ht]
\centering
\caption{Synthetic data evaluation for WESAD. For each column, best performance are in bold. The \textit{Sample} group represents average sample-level pairwise distance metrics, while the \textit{Distribution} group includes metrics computed at the distribution level. All metrics are reported as averages across classes.}
\label{tab:wesad_metrics}
\resizebox{\textwidth}{!}{
\begin{tabular}{|c|c|c|c|c|c|c|c|c|c|c|c|c|c|c|c|c|}
\hline
& & &\multicolumn{7}{c|}{\textbf{ECG Quality}}&\multicolumn{7}{c|}{\textbf{EDA Quality}}\\
\hline
& & &\multicolumn{3}{c|}{\textbf{Sample (features)}}&\textbf{Sample (raw)} &\multicolumn{3}{c|}{\textbf{Distribution}}&\multicolumn{3}{c|}{\textbf{Sample (features)}}&\textbf{Sample (raw)} &\multicolumn{3}{c|}{\textbf{Distribution}}\\
\cline{4-10} \cline{11-17}
\textbf{Model}& \textbf{Config ID} & \textbf{Best} &\textbf{CD} & \textbf{CrD}& \textbf{L2}& \textbf{DTWD} & \textbf{MMD} & \textbf{E} & \textbf{DS}& \textbf{CD} & \textbf{CrD}& \textbf{L2}& \textbf{DTWD} & \textbf{MMD} & \textbf{En} & \textbf{DS}\\ \hline  
TTS-CGAN& N-B8 & EDA &0.13& 0.09&1.04 & 11.1& 1.99& 0.42& 0.49& 0.12& 0.25& 6.17& 86.3&1.2& 594.4& 0.39\\ \hline
TTS-CGAN& U-B16 & ECG &0.03& \textbf{0.02}& 0.76& \textbf{9.65}& 1.98&\textbf{0.41}& 0.49& 0.64& 0.64& 5.24& 73.8& 1.36& 596.9& 0.43\\ \hline
P$2$P$_{COND}$& N-B16& ECG, EDA & 0.78 & 0.68& 4.46 &55.7& 1.98& 497.1& 0.49& 0.47& 0.83& 32.7&161.4& 1.68& 538.4&0.44\\ \hline
WaveGAN$^*$$_{COND}$ & N-B8 & ECG, EDA  & 0.75& 0.62& 4.1& 44.3& 1.98& 497.6& 0.49& 0.4& 0.74& 16.1& 114.2& 1.2& \textbf{537.2}& 0.41\\ \hline
BioDiffusion& N-B8& ECG, EDA  & \textbf{0.02}& \textbf{0.02}& \textbf{0.55}& 10.8& \textbf{1.88}& 0.71& \textbf{0.28}& \textbf{0.004}& \textbf{0.01}
& \textbf{4.4}&\textbf{62.5}& \textbf{0.87}& 596.8& \textbf{0.17}\\ \hline
SSSD & U-B32 &ECG, EDA  & 0.03
& \textbf{0.02}& 0.9& 13.2& 1.99& 0.79& 0.46& 0.10& 0.32& 6.3& 72.1& 1.67& 596.0&0.43\\ \hline
\end{tabular}}
\end{table*}

\begin{table*}[ht]
\centering
\caption{Synthetic data evaluation for SWELL. For each column, best performance are in bold.}
\label{tab:swell_metrics}
\resizebox{\textwidth}{!}{
\begin{tabular}{|c|c|c|c|c|c|c|c|c|c|c|c|c|c|c|c|c|}
\hline
& & &\multicolumn{7}{c|}{\textbf{ECG Quality}}&\multicolumn{7}{c|}{\textbf{EDA Quality}}\\
\hline
& & &\multicolumn{3}{c|}{\textbf{Sample (features)}}&\textbf{Sample (raw)} &\multicolumn{3}{c|}{\textbf{Distribution}}&\multicolumn{3}{c|}{\textbf{Sample (features)}}&\textbf{Sample (raw)} &\multicolumn{3}{c|}{\textbf{Distribution}}\\
\cline{4-10} \cline{11-17}
\textbf{Model}& \textbf{Config} & \textbf{Best} &\textbf{CD} & \textbf{CrD}& \textbf{L2}& \textbf{DTWD} & \textbf{MMD} & \textbf{E} & \textbf{DS}& \textbf{CD} & \textbf{CrD}& \textbf{L2}& \textbf{DTWD} & \textbf{MMD} & \textbf{En} & \textbf{DS}\\ \hline  
TTS-CGAN& U-B16& EDA &0.91&0.83& 51551.3&14179.5&1.0&4.3&0.49&0.05&0.18&930.7&8816.2&\textbf{0.5}&669.7&0.38\\ \hline
TTS-CGAN& N-B16& ECG&\textbf{0.003}&\textbf{0.002}&\textbf{10160.7}&\textbf{7300.5}&\textbf{0.5}&\textbf{3.9}&0.47& 0.87&1.46&28707.4&9766.0&1.0&669.5&0.44\\ \hline
P$2$P$_{COND}$& N-B8& ECG, EDA &0.03&0.02&49427.1&6748.0&1.0&639.4&0.49& 1.23&1.4&528617&58221.1&1.0&\textbf{648.1}&0.49\\ \hline
WaveGAN$^*$$_{COND}$& N-B16& ECG, EDA& 0.04&0.05&50560.4&6879.4&1.0&641.5&0.45&1.23&1.4&528617&88057.4&1.0&\textbf{648.1}&0.46\\ \hline
BioDiffusion& N-B32& ECG, EDA& 0.01&0.02&37804.1&7808.6&1.1&\textbf{3.9}&\textbf{0.39}&\textbf{0.04}&\textbf{0.11}&\textbf{578.0}&\textbf{5799.4}&\textbf{0.5}&669.1&\textbf{0.16}\\ \hline
SSSD& N-B32& EDA&0.97&1.15&51556.2&6952.2&1.35&4.0&0.49&\textbf{0.04}&0.12&651.9& 7790.3&0.78&669.4&0.44\\ \hline
SSSD&U-B32& ECG&0.50&0.65&51532.6&6880.9&1.0&3.8 &0.42&0.50&1.22&677.9&8091.2&1.0&669.6&0.47\\ \hline
\end{tabular}}
\end{table*}

\begin{table*}[ht]
\centering
\caption{Synthetic data evaluation for CASE. For each column, best performance are in bold.}
\label{tab:case_metrics}
\resizebox{\textwidth}{!}{
\begin{tabular}{|c|c|c|c|c|c|c|c|c|c|c|c|c|c|c|c|c|}
\hline
& & &\multicolumn{7}{c|}{\textbf{ECG Quality}}&\multicolumn{7}{c|}{\textbf{EDA Quality}}\\
\hline
& & &\multicolumn{3}{c|}{\textbf{Sample (features)}}&\textbf{Sample (raw)} &\multicolumn{3}{c|}{\textbf{Distribution}}&\multicolumn{3}{c|}{\textbf{Sample (features)}}&\textbf{Sample (raw)} &\multicolumn{3}{c|}{\textbf{Distribution}}\\
\cline{4-10} \cline{11-17}
\textbf{Model}& \textbf{Config} & \textbf{Best} &\textbf{CD} & \textbf{CrD}& \textbf{L2}& \textbf{DTWD} & \textbf{MMD} & \textbf{E} & \textbf{DS}& \textbf{CD} & \textbf{CrD}& \textbf{L2}& \textbf{DTWD} & \textbf{MMD} & \textbf{En} & \textbf{DS}\\ \hline  
TTS-CGAN& U-B32& ECG, EDA& 0.05&0.09& 1.12& 13.1& 1.77&\textbf{1.69}
& 0.49& 0.001& 0.004& 25.1& 353.7&\textbf{1.0}&312.2&0.34\\ \hline
P$2$P$_{COND}$& N-B16& ECG, EDA& 1.18& 1.22& 49.5&323.4& 1.96&478.3& 0.49& 1.28& 1.51&554.3&1144.8&1.5&486.1&0.48\\ \hline
WaveGAN$^*$$_{COND}$& N-B16& ECG, EDA&1.09& 0.91 &18.5 &254.6 &1.93 &477.6& 0.49&1.17&1.43&84.8&777.5&\textbf{1.0}&482.1&0.49\\ \hline
BioDiffusion& N-B8& ECG, EDA& \textbf{0.04}& 0.08& \textbf{0.76}& \textbf{10.8}&\textbf{1.47}& 1.84& \textbf{0.23}& \textbf{0.0006}&\textbf{0.0002}&\textbf{22.8}&\textbf{313.5}&\textbf{1.0}&\textbf{309.7}&\textbf{0.03}\\ \hline
SSSD& N-B32& EDA& 0.14& 0.30& 1.67&13.7&1 .96&2.79& 0.49&0.45&0.75&41.7&550.4&\textbf{1.0}& 314.0&0.40\\ \hline
SSSD& U-B32& ECG&\textbf{0.04}&\textbf{0.06}&1.13&11.4&1.96&1.86&0.45& 0.60&0.89&47.8&674.5&1.5&314.7& 0.45\\ \hline
\end{tabular}}
\end{table*}

\subsection{Model comparison}
\label{sec:model_comparison}
Tables \ref{tab:wesad_metrics} to \ref{tab:case_metrics} present the evaluation of synthetic TS data quality. The primary goal is to identify the top-performing configuration for each model that achieves the highest quality for both ECG and EDA signals simultaneously. In case this is not possible, the best-performing configurations for each modality are reported separately. On the other hand, Table \ref{tab:utility} shows the best performance of synthetic data in downstream tasks. As it may be expected, configurations that yield the best quality for both signals also correspond to the highest utility. However, when the optimal configuration differs between the two signals, it becomes necessary to identify the best trade-off between similarity and downstream task performance.
\\
A clear pattern emerges across all datasets, with BioDiffusion consistently outperforming the other models. This is particularly evident for WESAD, where BioDiffusion achieves the lowest values for almost all metrics for both ECG and EDA signals. TTS-CGAN provides competitive results, marginally surpassing BioDiffusion in DTWD (temporal modeling) and data entropy (diversity), with SSSD following closely. A similar trend is observed for CASE, where SSSD emerges as the runner-up model. For SWELL, BioDiffusion shows superior performance in EDA, while TTS-CGAN excels in ECG modeling. However, BioDiffusion exhibits a lower discriminative score for ECG and it achieves the smallest drop in TSTR performance for all datasets, while leading to the highest improvements in average DA scores.
Figure \ref{fig:tsne} shows the overlap between real and synthetic distributions for each signal modality and class, based on the top-performing BioDiffusion configuration for each dataset. These visualizations highlight satisfactory coverage of real data, as well as considerable diversity within the synthetic distributions for nearly all signal-class pairs.
\\
Following BioDiffusion, TTS-CGAN and SSSD exhibit dataset-specific strengths: TTS-CGAN performs better on the WESAD and SWELL datasets, while SSSD excels on CASE.
Finally, P$2$P$_{COND}$ and WaveGAN$^*$$_{COND}$ consistently underperform, also exhibiting class-specific mode collapse. As a result, quantitative assessment further emphasizes their comparable behavior in terms of data quality and utility metrics. This can likely be attributed to several shared features, such as identical discriminator, batch normalization for conditioning, and deconvolution-based generator networks (despite with different architectures).

\subsection{Signal quality assessment}
\label{res:quality}
Analysis of synthetic signal quality indicates that joint multimodal generation remains suboptimal.
For TTS-CGAN and SSSD, configurations optimized for ECG fail to produce comparable results for EDA, and vice versa. Even in cases where a \textit{\quotes{win-win}} scenario is achieved, substantial performance gaps between modalities persist. Results also indicate that ECG is the most challenging signal to replicate due to its intricate waveforms, whereas EDA, with its slow-varying dynamics and event-related peaks, yields better results.
Focusing on the best-performing BioDiffusion model, distribution-level metrics such as MMD and discriminative scores reveal notable differences between ECG and EDA. Specifically, the MMD differences are $+1.01$, $+0.6$, and $+0.47$, while the discriminative scores show increases of $+11\%$, $+23\%$, and $+20\%$ for WESAD, SWELL, and CASE datasets, respectively. These results underscore the lower realism of synthetic ECG compared to EDA. MMD values for ECG are near the upper bound for both WESAD and CASE, while for SWELL, they range between $1$ and $1.35$, except for TTS-CGAN (MMD = $0.5$). For all models except BioDiffusion, discriminative scores for ECG exceed $40\%$ and approach the upper bound in most instances. Moreover, our findings suggest that ECG is the most influential feature for downstream predictive tasks, as configurations of TTS-CGAN and SSSD optimized for ECG quality consistently yield the best data utility, highlighting its crucial role in predictive performance.
\\
Regarding LRD modeling, DTWD is the most suitable metric, as it is specifically designed to assess similarity in temporal dynamics. Results show that BioDiffusion achieves the most effective modeling in $4$ out of $6$ dataset-signal pairs, while TTS-CGAN slightly outperforms BioDiffusion in the remaining two cases (ECG signals from WESAD and SWELL). However, since DTWD is unbounded and data-dependent, direct performance comparison between different signals is not feasible. Furthermore, both BioDiffusion and TTS-CGAN share the use of attention mechanism, reflecting established findings on its strength in capturing LRD despite scalability challenges.
\\
Finally, it should be noticed that feature-based distance metrics (CD, CrD, and L$2$) often show minimal differences among top models. These metrics mainly reflect alignment in statistical properties and can overestimate synthetic data quality when used alone, overlooking key aspects such as temporal or morphological fidelity. In contrast, distribution-level metrics such as MMD and discriminative scores reveal more pronounced differences across modalities and models that correlate more closely with data utility assessment, as they align with poor performance in downstream tasks even if feature-based metrics appear favorable.

\begin{table*}[ht]
\centering
\caption{Synthetic data utility.} 
\label{tab:utility}
\begin{tabular}{|c|c|c|c|c|c|c|c|c|c|}
\hline
& \multicolumn{3}{c|}{\textbf{WESAD}}&\multicolumn{3}{c|}{\textbf{SWELL}}&\multicolumn{3}{c|}{\textbf{CASE}}\\
\cline{2-4} \cline{5-7}\cline{8-10}
\textbf{Model}& \textbf{Config ID} & \textbf{TSTR}& \textbf{DA}& \textbf{Config ID} & \textbf{TSTR}& \textbf{DA}& \textbf{Config ID} & \textbf{TSTR}& \textbf{DA}\\ \hline
TTS-CGAN & U-B16& -21.4& +0.2 &N-B16 &-9.3 &+1.5 &U-B32 &-3.9 &+1.5 \\ \hline
P$2$P$_{COND}$& N-B16&-25.8& -0.7 &N-B8 &-14.6 &+1.1 &N-B16 &-5.7 &-2.2 \\ \hline
WaveGAN$^*$$_{COND}$ & N-B8& -22.5& -2.0&N-B16 &-16.0 &+0.9 &N-B16 &-6.2 &-1.8 \\
\hline
BioDiffusion& N-B8&\textbf{-3.5}&\textbf{+2.5}&N-B32&\textbf{-6.2}&\textbf{+3.1}&N-B8&\textbf{-0.7}&\textbf{+1.9}\\ 
\hline
SSSD& U-B32& -13.5&-2.3&U-B32&-15.4&-3.2&U-B32&-13.5&-2.3\\
\hline
\end{tabular}
\end{table*}

\subsection{Data utility assessment}
The results presented in Table \ref{tab:utility} clearly indicates that training solely with synthetic data consistently results in a decrease in predictive performance. As already mentioned in Section \ref{sec:model_comparison}, BioDiffusion achieves the smallest drop across all datasets, ranging from $-0.7\%$ (CASE), $-3.5\%$ (WESAD), to $-6.2\%$ (SWELL).
Notably, the gap in TSTR scores between BioDiffusion and the runner-up model is significant for WESAD ($-10\%$), while it narrows for SWELL and CASE ($\approx3\%$). Consequently, training with synthetic data while maintaining an acceptable performance is feasible primarily for BioDiffusion and, in few cases, for other models (e.g., TTS-CGAN on the CASE dataset).
\\
Conversely, the improvements in DA performance remain modest, with BioDiffusion achieving average gains of $2$–$3\%$ in the best case, while the other models exhibit negligible enhancements or even slight declines.
It is important to highlight that our selected DA policies imply a limited synthetic-to-real data ratio, which is $\leq1$ for the \textit{Balance} and \textit{Double} settings, with only the \textit{Balance$+$Double} settings requiring a substantial amount of synthetic data (ratio $>1$) Therefore, DA performance may deteriorate as more synthetic data is incorporated into training unless stronger guarantees of data quality are ensured. Furthermore, the modest improvements observed in DA highlight the need for novel metrics to assess the generalizability and novelty of synthetic data. Ideally, synthetic samples should balance sufficient divergence to expand the understanding of class distributions with adequate alignment to the original source distribution to enhance predictive outcomes effectively.

\subsection{Insights and limitations of synthetic TS evaluation}
\label{sec:limitations}
generation).
The major practical limitation of our study lies in selecting the best model checkpoint for inference (i.e., data generation).
As described in Section \ref{subsec:config}, checkpoints were chosen post-training based on their respective loss functions. This strategy, adapted from conventional DL practices, differs from common approaches in generative AI for CV, where evaluation and inference are often integrated into the training loop. In those cases, a subset of generated images—initialized from a fixed latent noise—is periodically monitored to guide model selection. For DDPM, our approach is motivated by the high computational cost of inference inherent to the diffusion process, particularly in low-resource settings, whereas GAN inference is comparatively faster. However, the absence of a universally accepted evaluation metric for TS data makes post-hoc model selection especially challenging.
\\
Despite its practicality, our approach has limitations. In GAN, an effective generator (low $G$ loss) must coexist with a strong discriminator (low $D$ loss); otherwise, poor-quality or repetitive outputs may go undetected. Therefore, we implemented a procedure to adjust checkpoint selection when needed, seeking $G$ minima that align with $D$ minima to ensure this balance, even if it required choosing a local $G$ optimum.
For DDPM, reconstruction losses exhibit an asymptotic trend that may suggest effective learning and convergence to a global minimum. However, losses are averaged across all $T=1000$ denoising timesteps. At higher timesteps, where the data are mostly noise, predictions can produce deceptively low errors. In contrast, early timesteps require precise reconstruction of real data, and averaging across all timesteps can mask suboptimal performance during this critical phase.
\\
These observations highlight the urgent need for standardized evaluation metrics, analogous to FID for synthetic images, to seamlessly integrate inference within training and enable more reliable model selection.
Recent proposals, such as Context-FID \cite{psa-gan} and Fréchet Transformer Distance (FTD) \cite{gat-gan}, attempt to replicate FID in the TS domain but lack the core strengths that make FID a trustworthy measure of image quality. In fact, FID captures both realism and diversity by comparing distributions in a perceptually meaningful embedding space derived from features aligned with human visual perception (e.g., shapes, textures).
In contrast, Context-FID and FTD rely on embeddings from TS foundation models (e.g., TS$2$Vec \cite{franceschi2019unsupervised}), whose temporal representations remain poorly understood. Additionally, these models are often trained on heterogeneous domains, limiting their transferability to specific contexts such as wearable sensor data, even after fine-tuning. 
Alternatively, downstream proxy metrics (e.g., TSTR) provide indirect insights into synthetic data quality and impose substantial computational overhead during training. Although several TS classifiers exist, they generally require task-specific fine-tuning or even full retraining, also limiting their generalizability as universal evaluators.

\section{Conclusions and Future works}
\label{sec:future_works}
This study provides a comprehensive evaluation of generative AI frameworks for wearable sensor data, which typically consist of multiple heterogeneous TS streams. Generating such data poses several challenges: producing joint and coherent multimodal outputs, synthesizing sequences long enough for meaningful downstream inference, and incorporating metadata to condition the generation, allowing customization at different levels.
Our experiments demonstrate that SoTA TS generative models face significant limitations in this complex task. As a preliminary outcome, training multiple class-specific unconditional models (TimeGAN) results in extremely low-quality data, while more meaningful outcomes are obtained through conditional generation. Nevertheless, most generated signal pairs remain suboptimal: some configurations unfairly prioritize one modality over the other, and even \quotes{win-win} setups exhibit substantial cross-modal disparities across evaluation metrics.
For LRD modeling, incorporating Transformers and/or attention mechanisms consistently improves long-term pattern synthesis, as reflected by lower DTWD values, though at higher computational cost. Among the models tested, only BioDiffusion achieves consistently satisfactory quality across modalities, with also minimal TSTR degradation. However, improvements in DA are modest, limiting the practical applicability of the generated data.
\\
Building on our findings, we plan to address these challenges from different perspectives.
For multimodal TS generation, we are analyzing more recent multi-agent GAN frameworks \cite{coscigan2022, chen2023hmgan, desmet2024hydra}, which balance intra- and cross-modal realism through multiple or hierarchical discriminators, thus enabling multi-objective optimization.
However, since temporal coherence and spectral consistency are fundamental TS properties beyond realism, we are developing a TS-centric, multi-task objective with adaptive weighting to jointly and automatically optimize multiple components. This framework can be applied to both GANs and DDPM, extending standard adversarial or reconstruction losses by incorporating specialized objectives such as DTW, autocorrelation, and Fourier-based terms.
Moreover, Mixture-of-Experts (MoE) Transformer architectures \cite{chen2024pathformer}, operating on different patch sizes, can naturally handle sequential data while supporting multi-scale temporal modeling across various resolutions and distances. This addresses a key limitation of current TS generative models, which often struggle to capture both short- and long-term dependencies. Performance can be further enhanced by developing robust multimodal representation learning methods, such as combining specialized attention modules to balance cross-modal fusion and intra-modal refinement \cite{zhang2023crossformer}, and using causal masks to model more realistic temporal transitions \cite{liu2025timerxl}.
\\
The final major challenge is personalization. Incorporating structured metadata as contextual information can enable community-level data generation, which is particularly valuable for human sensing applications targeting underrepresented or \quotes{hard-to-reach} populations (e.g., older adults).
However, this introduces challenges in handling high-dimensional conditioning spaces, requiring models to learn sparse input-output relationships. To address this, we plan to investigate strategies for simplifying the conditioning space (e.g., via low-level embeddings) and developing interpolation techniques to improve coverage across similar distributions. Additionally, intra- and inter-user data translation may enable subject-level personalization; we will investigate cyclic GAN architectures, commonly used for unpaired domain translation, to adapt multimodal data to specific user characteristics and application domains.
\\
Another timely and relevant direction is text-conditioned generation, which could enrich sensor data with semantic information. However, identifying meaningful textual descriptions is challenging and often restricted to clinical settings (e.g., ECG reports \cite{chung2023autotte}) or specific tasks, such as HAR \cite{leng2023generating}. Very recent generalist approaches \cite{gu2025verbalts} are limited to a small set of attributes, such as trend and seasonality, which may be absent or irrelevant for wearable sensor data. Consequently, generating informative and contextually relevant text for these signals remains an open challenge for improving wearable data synthesis.

\bibliographystyle{ieeetr}
\bibliography{refs.bib}

@article{alcaraz2023sssd_ecg,
  title={Diffusion-based conditional ECG generation with structured state space models},
  author={Alcaraz, Juan Miguel Lopez and Strodthoff, Nils},
  journal={Computers in biology and medicine},
  volume={163},
  pages={107115},
  year={2023},
  publisher={Elsevier}
}

@inproceedings{peebles2023scalable,
  title={Scalable diffusion models with transformers},
  author={Peebles, William and Xie, Saining},
  booktitle={Proceedings of the IEEE/CVF International Conference on Computer Vision},
  pages={4195--4205},
  year={2023}
}

@article{sikder2025transfusion,
  title={Transfusion: generating long, high fidelity time series using diffusion models with transformers},
  author={Sikder, Md Fahim and Ramachandranpillai, Resmi and Heintz, Fredrik},
  journal={Machine Learning with Applications},
  volume={20},
  pages={100652},
  year={2025},
  publisher={Elsevier}
}

@article{esteban2017real,
  title={Real-valued (medical) time series generation with recurrent conditional gans},
  author={Esteban, Crist{\'o}bal and Hyland, Stephanie L and R{\"a}tsch, Gunnar},
  journal={arXiv preprint arXiv:1706.02633},
  year={2017}
}

@article{brophy2023generative,
  title={Generative adversarial networks in time series: A systematic literature review},
  author={Brophy, Eoin and Wang, Zhengwei and She, Qi and Ward, Tom{\'a}s},
  journal={ACM Computing Surveys},
  volume={55},
  number={10},
  pages={1--31},
  year={2023},
  publisher={ACM New York, NY}
}

@article{yoon2019time,
  title={Time-series generative adversarial networks},
  author={Yoon, Jinsung and Jarrett, Daniel and Van der Schaar, Mihaela},
  journal={Advances in neural information processing systems},
  volume={32},
  year={2019}
}

@inproceedings{
donahueadversarial,
title={Adversarial Audio Synthesis},
author={Chris Donahue and Julian McAuley and Miller Puckette},
booktitle={International Conference on Learning Representations},
year={2019}
}

@inproceedings{li2022ttsgan,
  title={Tts-gan: A transformer-based time-series generative adversarial network},
  author={Li, Xiaomin and Metsis, Vangelis and Wang, Huangyingrui and Ngu, Anne Hee Hiong},
  booktitle={International conference on artificial intelligence in medicine},
  pages={133--143},
  year={2022},
  organization={Springer}
}

@article{li2022ttscgan,
  title={Tts-cgan: A transformer time-series conditional gan for biosignal data augmentation},
  author={Li, Xiaomin and Ngu, Anne Hee Hiong and Metsis, Vangelis},
  journal={arXiv preprint arXiv:2206.13676},
  year={2022}
}

@article{radford2015unsupervised,
  title={Unsupervised representation learning with deep convolutional generative adversarial networks},
  author={Radford, Alec},
  journal={arXiv preprint arXiv:1511.06434},
  year={2015}
}

@article{thambawita2021deepfake,
  title={DeepFake electrocardiograms using generative adversarial networks are the beginning of the end for privacy issues in medicine},
  author={Thambawita, Vajira and Isaksen, Jonas L and Hicks, Steven A and Ghouse, Jonas and Ahlberg, Gustav and Linneberg, Allan and Grarup, Niels and Ellervik, Christina and Olesen, Morten Salling and Hansen, Torben and others},
  journal={Scientific reports},
  volume={11},
  number={1},
  pages={21896},
  year={2021},
  publisher={Nature Publishing Group UK London}
}

@inproceedings{
dosovitskiy2021an,
title={An Image is Worth 16x16 Words: Transformers for Image Recognition at Scale},
author={Alexey Dosovitskiy and Lucas Beyer and Alexander Kolesnikov and Dirk Weissenborn and Xiaohua Zhai and Thomas Unterthiner and Mostafa Dehghani and Matthias Minderer and Georg Heigold and Sylvain Gelly and Jakob Uszkoreit and Neil Houlsby},
booktitle={International Conference on Learning Representations},
year={2021}
}

@inproceedings{sohl2015deep,
  title={Deep unsupervised learning using nonequilibrium thermodynamics},
  author={Sohl-Dickstein, Jascha and Weiss, Eric and Maheswaranathan, Niru and Ganguli, Surya},
  booktitle={International conference on machine learning},
  pages={2256--2265},
  year={2015},
  organization={PMLR}
}

@article{ho2020denoising,
  title={Denoising diffusion probabilistic models},
  author={Ho, Jonathan and Jain, Ajay and Abbeel, Pieter},
  journal={Advances in neural information processing systems},
  volume={33},
  pages={6840--6851},
  year={2020}
}

@article{dhariwal2021diffusion,
  title={Diffusion models beat gans on image synthesis},
  author={Dhariwal, Prafulla and Nichol, Alexander},
  journal={Advances in neural information processing systems},
  volume={34},
  pages={8780--8794},
  year={2021}
}

@article{kang2022augmented,
  title={Augmented adversarial learning for human activity recognition with partial sensor sets},
  author={Kang, Hua and Huang, Qianyi and Zhang, Qian},
  journal={Proceedings of the ACM on Interactive, Mobile, Wearable and Ubiquitous Technologies},
  volume={6},
  number={3},
  pages={1--30},
  year={2022},
  publisher={ACM New York, NY, USA}
}

@article{chen2023hmgan,
  title={HMGAN: A hierarchical multi-modal generative adversarial network model for wearable human activity recognition},
  author={Chen, Ling and Hu, Rong and Wu, Menghan and Zhou, Xin},
  journal={Proceedings of the ACM on Interactive, Mobile, Wearable and Ubiquitous Technologies},
  volume={7},
  number={3},
  pages={1--27},
  year={2023},
  publisher={ACM New York, NY, USA}
}

@article{desmet2024hydra,
  title={Hydra-TS: Enhancing Human Activity Recognition with Multi-Objective Synthetic Time Series Data Generation},
  author={DeSmet, Chance and Greeley, Colin and Cook, Diane J},
  journal={IEEE Sensors Journal},
  year={2024},
  publisher={IEEE}
}

@article{li2024biodiffusion,
  title={Biodiffusion: A versatile diffusion model for biomedical signal synthesis},
  author={Li, Xiaomin and Sakevych, Mykhailo and Atkinson, Gentry and Metsis, Vangelis},
  journal={Bioengineering},
  volume={11},
  number={4},
  pages={299},
  year={2024},
  publisher={MDPI}
}

@inproceedings{gu2022efficiently,
title={Efficiently Modeling Long Sequences with Structured State Spaces},
author={Albert Gu and Karan Goel and Christopher Re},
booktitle={International Conference on Learning Representations},
year={2022},
}

@article{coscigan2022,
  title={Generating multivariate time series with COmmon Source CoordInated GAN (COSCI-GAN)},
  author={Seyfi, Ali and Rajotte, Jean-Francois and Ng, Raymond},
  journal={Advances in neural information processing systems},
  volume={35},
  pages={32777--32788},
  year={2022}
}

@article{li2023descod,
  title={Descod-ecg: Deep score-based diffusion model for ecg baseline wander and noise removal},
  author={Li, Huayu and Ditzler, Gregory and Roveda, Janet and Li, Ao},
  journal={IEEE Journal of Biomedical and Health Informatics},
  year={2023},
  publisher={IEEE}
}

@article{zhu2019electrocardiogram,
  title={Electrocardiogram generation with a bidirectional LSTM-CNN generative adversarial network},
  author={Zhu, Fei and Ye, Fei and Fu, Yuchen and Liu, Quan and Shen, Bairong},
  journal={Scientific reports},
  volume={9},
  number={1},
  pages={6734},
  year={2019},
  publisher={Nature Publishing Group UK London}
}

@inproceedings{sharma2023medic,
  title={Medic: Mitigating EEG data scarcity via class-conditioned diffusion model},
  author={Sharma, Gulshan and Dhall, Abhinav and Subramanian, Ramanathan},
  booktitle={Deep Generative Models for Health Workshop NeurIPS 2023},
  year={2023}
}

@article{kiyasseh2020plethaugment,
  title={PlethAugment: GAN-based PPG augmentation for medical diagnosis in low-resource settings},
  author={Kiyasseh, Dani and Tadesse, Girmaw Abebe and Thwaites, Louise and Zhu, Tingting and Clifton, David and others},
  journal={IEEE journal of biomedical and health informatics},
  volume={24},
  number={11},
  pages={3226--3235},
  year={2020},
  publisher={IEEE}
}

@article{wang2022wearable,
  title={A wearable-HAR oriented sensory data generation method based on spatio-temporal reinforced conditional GANs},
  author={Wang, Jiwei and Chen, Yiqiang and Gu, Yang},
  journal={Neurocomputing},
  volume={493},
  pages={548--567},
  year={2022},
  publisher={Elsevier}
}

@article{ehrhart2022conditional,
  title={A conditional gan for generating time series data for stress detection in wearable physiological sensor data},
  author={Ehrhart, Maximilian and Resch, Bernd and Havas, Clemens and Niederseer, David},
  journal={Sensors},
  volume={22},
  number={16},
  pages={5969},
  year={2022},
  publisher={MDPI}
}

@article{kumar2019melgan,
  title={Melgan: Generative adversarial networks for conditional waveform synthesis},
  author={Kumar, Kundan and Kumar, Rithesh and De Boissiere, Thibault and Gestin, Lucas and Teoh, Wei Zhen and Sotelo, Jose and De Brebisson, Alexandre and Bengio, Yoshua and Courville, Aaron C},
  journal={Advances in neural information processing systems},
  volume={32},
  year={2019}
}

@inproceedings{psa-gan,
  title={PSA-GAN: Progressive Self Attention GANs for Synthetic Time Series},
  author={Jeha, Paul and Bohlke-Schneider, Michael and Mercado, Pedro and Kapoor, Shubham and Nirwan, Rajbir Singh and Flunkert, Valentin and Gasthaus, Jan and Januschowski, Tim},
  year={2022},
  booktitle={International Conference on Learning Representations}
}

@article{gat-gan,
  title={GAT-GAN: A Graph-Attention-based Time-Series Generative Adversarial Network},
  author={Iyer, Srikrishna and Hou, Teng Teck},
  journal={arXiv preprint arXiv:2306.01999},
  year={2023}
}

@article{heusel2017gans,
  title={Gans trained by a two time-scale update rule converge to a local nash equilibrium},
  author={Heusel, Martin and Ramsauer, Hubert and Unterthiner, Thomas and Nessler, Bernhard and Hochreiter, Sepp},
  journal={Advances in neural information processing systems},
  volume={30},
  year={2017}
}

@article{timevae,
  title={Timevae: A variational auto-encoder for multivariate time series generation},
  author={Desai, Abhyuday and Freeman, Cynthia and Wang, Zuhui and Beaver, Ian},
  journal={arXiv preprint arXiv:2111.08095},
  year={2021}
}

@article{normalizingflows,
  title={Normalizing flows: An introduction and review of current methods},
  author={Kobyzev, Ivan and Prince, Simon JD and Brubaker, Marcus A},
  journal={IEEE transactions on pattern analysis and machine intelligence},
  volume={43},
  number={11},
  pages={3964--3979},
  year={2020},
  publisher={IEEE}
}

@inproceedings{ebm,
  title={Learning deep energy models},
  author={Ngiam, Jiquan and Chen, Zhenghao and Koh, Pang W and Ng, Andrew Y},
  booktitle={Proceedings of the 28th international conference on machine learning (ICML-11)},
  pages={1105--1112},
  year={2011}
}

@inproceedings{emotionalgan,
  title={EmotionalGAN: Generating ECG to enhance emotion state classification},
  author={Chen, Genlang and Zhu, Yi and Hong, Zhiqing and Yang, Zhen},
  booktitle={Proceedings of the 2019 International conference on artificial intelligence and computer science},
  pages={309--313},
  year={2019}
}

@inproceedings{unet,
  title={U-net: Convolutional networks for biomedical image segmentation},
  author={Ronneberger, Olaf and Fischer, Philipp and Brox, Thomas},
  booktitle={Medical image computing and computer-assisted intervention--MICCAI 2015: 18th international conference, Munich, Germany, October 5-9, 2015, proceedings, part III 18},
  pages={234--241},
  year={2015},
  organization={Springer}
}

@article{de2017modulating,
  title={Modulating early visual processing by language},
  author={De Vries, Harm and Strub, Florian and Mary, J{\'e}r{\'e}mie and Larochelle, Hugo and Pietquin, Olivier and Courville, Aaron C},
  journal={Advances in neural information processing systems},
  volume={30},
  year={2017}
}

@article{vaswani2017attention,
  title={Attention is all you need},
  author={Vaswani, A},
  journal={Advances in Neural Information Processing Systems},
  year={2017}
}

@inproceedings{swell,
  title={The swell knowledge work dataset for stress and user modeling research},
  author={Koldijk, Saskia and Sappelli, Maya and Verberne, Suzan and Neerincx, Mark A and Kraaij, Wessel},
  booktitle={Proceedings of the 16th international conference on multimodal interaction},
  pages={291--298},
  year={2014}
}

@inproceedings{wesad,
  title={Introducing wesad, a multimodal dataset for wearable stress and affect detection},
  author={Schmidt, Philip and Reiss, Attila and Duerichen, Robert and Marberger, Claus and Van Laerhoven, Kristof},
  booktitle={Proceedings of the 20th ACM international conference on multimodal interaction},
  pages={400--408},
  year={2018}
}

@article{case,
  title={A dataset of continuous affect annotations and physiological signals for emotion analysis},
  author={Sharma, Karan and Castellini, Claudio and Van Den Broek, Egon L and Albu-Schaeffer, Alin and Schwenker, Friedhelm},
  journal={Scientific data},
  volume={6},
  number={1},
  pages={196},
  year={2019},
  publisher={Nature Publishing Group UK London}
}

@article{goodfellow2014generative,
  title={Generative adversarial nets},
  author={Goodfellow, Ian and Pouget-Abadie, Jean and Mirza, Mehdi and Xu, Bing and Warde-Farley, David and Ozair, Sherjil and Courville, Aaron and Bengio, Yoshua},
  journal={Advances in neural information processing systems},
  volume={27},
  year={2014}
}

@article{oord2016wavenet,
  title={WaveNet: A Generative Model for Raw Audio},
  author={Oord, Aaron van den},
  journal={arXiv preprint arXiv:1609.03499},
  year={2016}
}

@inproceedings{ronneberger2015u,
  title={U-net: Convolutional networks for biomedical image segmentation},
  author={Ronneberger, Olaf and Fischer, Philipp and Brox, Thomas},
  booktitle={Medical image computing and computer-assisted intervention--MICCAI 2015: 18th international conference, Munich, Germany, October 5-9, 2015, proceedings, part III 18},
  pages={234--241},
  year={2015},
  organization={Springer}
}

@article{gulrajani2017improved,
  title={Improved training of wasserstein gans},
  author={Gulrajani, Ishaan and Ahmed, Faruk and Arjovsky, Martin and Dumoulin, Vincent and Courville, Aaron C},
  journal={Advances in neural information processing systems},
  volume={30},
  year={2017}
}

@inproceedings{zhu2017unpaired,
  title={Unpaired image-to-image translation using cycle-consistent adversarial networks},
  author={Zhu, Jun-Yan and Park, Taesung and Isola, Phillip and Efros, Alexei A},
  booktitle={Proceedings of the IEEE international conference on computer vision},
  pages={2223--2232},
  year={2017}
}

@article{yang2024survey,
  title={A survey on diffusion models for time series and spatio-temporal data},
  author={Yang, Yiyuan and Jin, Ming and Wen, Haomin and Zhang, Chaoli and Liang, Yuxuan and Ma, Lintao and Wang, Yi and Liu, Chenghao and Yang, Bin and Xu, Zenglin and others},
  journal={arXiv preprint arXiv:2404.18886},
  year={2024}
}

@inproceedings{
kong2021diffwave,
title={DiffWave: A Versatile Diffusion Model for Audio Synthesis},
author={Zhifeng Kong and Wei Ping and Jiaji Huang and Kexin Zhao and Bryan Catanzaro},
booktitle={International Conference on Learning Representations},
year={2021},
}

@article{xiong2024patchemg,
  title={Patchemg: Few-shot emg signal generation with diffusion models for data augmentation to improve classification performance},
  author={Xiong, Baoping and Chen, Wensheng and Li, Han and Niu, Yinxi and Zeng, Nianyin and Gan, Zhenhua and Xu, Yong},
  journal={IEEE Transactions on Instrumentation and Measurement},
  year={2024},
  publisher={IEEE}
}

@inproceedings{aristimunha2023synthetic,
  title={Synthetic sleep EEG signal generation using latent diffusion models},
  author={Aristimunha, Bruno and de Camargo, Raphael Yokoingawa and Chevallier, Sylvain and Lucena, Oeslle and Thomas, Adam G and Cardoso, M Jorge and Pinaya, Walter Hugo Lopez and Dafflon, Jessica},
  booktitle={Deep Generative Models for Health Workshop NeurIPS 2023},
  year={2023}
}

@inproceedings{
yuan2024diffusionts,
title={Diffusion-{TS}: Interpretable Diffusion for General Time Series Generation},
author={Xinyu Yuan and Yan Qiao},
booktitle={The Twelfth International Conference on Learning Representations},
year={2024},
}

@inproceedings{rombach2022high,
  title={High-resolution image synthesis with latent diffusion models},
  author={Rombach, Robin and Blattmann, Andreas and Lorenz, Dominik and Esser, Patrick and Ommer, Bj{\"o}rn},
  booktitle={Proceedings of the IEEE/CVF conference on computer vision and pattern recognition},
  pages={10684--10695},
  year={2022}
}

@inproceedings{zhou2024mtsci,
  title={MTSCI: A Conditional Diffusion Model for Multivariate Time Series Consistent Imputation},
  author={Zhou, Jianping and Li, Junhao and Zheng, Guanjie and Wang, Xinbing and Zhou, Chenghu},
  booktitle={Proceedings of the 33rd ACM International Conference on Information and Knowledge Management},
  pages={3474--3483},
  year={2024}
}

@article{gao2023adversarial,
  title={Adversarial self-attentive time-variant neural networks for multi-step time series forecasting},
  author={Gao, Changxia and Zhang, Ning and Li, Youru and Lin, Yan and Wan, Huaiyu},
  journal={Expert Systems with Applications},
  volume={231},
  pages={120722},
  year={2023},
  publisher={Elsevier}
}

@inproceedings{
mehri2017samplernn,
title={Sample{RNN}: An Unconditional End-to-End Neural Audio Generation Model},
author={Soroush Mehri and Kundan Kumar and Ishaan Gulrajani and Rithesh Kumar and Shubham Jain and Jose Sotelo and Aaron Courville and Yoshua Bengio},
booktitle={International Conference on Learning Representations},
year={2017},
}

@inproceedings{goel2022sashimi,
  title={It’s raw! audio generation with state-space models},
  author={Goel, Karan and Gu, Albert and Donahue, Chris and R{\'e}, Christopher},
  booktitle={International Conference on Machine Learning},
  pages={7616--7633},
  year={2022},
  organization={PMLR}
}

@inproceedings{
choromanski2021rethinking,
title={Rethinking Attention with Performers},
author={Krzysztof Marcin Choromanski and Valerii Likhosherstov and David Dohan and Xingyou Song and Andreea Gane and Tamas Sarlos and Peter Hawkins and Jared Quincy Davis and Afroz Mohiuddin and Lukasz Kaiser and David Benjamin Belanger and Lucy J Colwell and Adrian Weller},
booktitle={International Conference on Learning Representations},
year={2021},
}

@inproceedings{odena2017acgan,  
  title={Conditional image synthesis with auxiliary classifier gans},
  author={Odena, Augustus and Olah, Christopher and Shlens, Jonathon},
  booktitle={International conference on machine learning},
  pages={2642--2651},
  year={2017},
  organization={PMLR}
}

@inproceedings{chung2023autotte,
  title={Text-to-ecg: 12-lead electrocardiogram synthesis conditioned on clinical text reports},
  author={Chung, Hyunseung and Kim, Jiho and Kwon, Joon-myoung and Jeon, Ki-Hyun and Lee, Min Sung and Choi, Edward},
  booktitle={ICASSP 2023-2023 IEEE International Conference on Acoustics, Speech and Signal Processing (ICASSP)},
  pages={1--5},
  year={2023},
  organization={IEEE}
}

@inproceedings{zhang2022mind,
  title={Mind Your Step: Continuous Conditional GANs with Generator Regularization},
  author={Zhang, Yunkai and Zheng, Yufeng and Ma, Xueying and Teng, Siyuan and Zheng, Zeyu},
  booktitle={NeurIPS 2022 Workshop on Synthetic Data for Empowering ML Research},
year= {2022}
}

@inproceedings{
ho2021classifierfree,
title={Classifier-Free Diffusion Guidance},
author={Jonathan Ho and Tim Salimans},
booktitle={NeurIPS 2021 Workshop on Deep Generative Models and Downstream Applications},
year={2021},
}

@article{stenger2024evaluation,
  title={Evaluation is key: a survey on evaluation measures for synthetic time series},
  author={Stenger, Michael and Leppich, Robert and Foster, Ian and Kounev, Samuel and Bauer, Andr{\'e}},
  journal={Journal of Big Data},
  volume={11},
  number={1},
  pages={66},
  year={2024},
  publisher={Springer}
}

@article{franceschi2019unsupervised,
  title={Unsupervised scalable representation learning for multivariate time series},
  author={Franceschi, Jean-Yves and Dieuleveut, Aymeric and Jaggi, Martin},
  journal={Advances in neural information processing systems},
  volume={32},
  year={2019}
}

@article{sharma2012objective,
  title={Objective measures, sensors and computational techniques for stress recognition and classification: A survey},
  author={Sharma, Nandita and Gedeon, Tom},
  journal={Computer methods and programs in biomedicine},
  volume={108},
  number={3},
  pages={1287--1301},
  year={2012},
  publisher={Elsevier}
}

@article{mehari2022advancing,
  title={Advancing the state-of-the-art for ECG analysis through structured state space models},
  author={Mehari, Temesgen and Strodthoff, Nils},
  journal={arXiv preprint arXiv:2211.07579},
  year={2022}
}

@article{society2012publication,
  title={Publication recommendations for electrodermal measurements},
  author={Society for Psychophysiological Research Ad Hoc Committee on Electrodermal Measures and Boucsein, Wolfram and Fowles, Don C and Grimnes, Sverre and Ben-Shakhar, Gershon and Roth, Walton T and Dawson, Michael E and Filion, Diane L},
  journal={Psychophysiology},
  volume={49},
  number={8},
  pages={1017--1034},
  year={2012},
  publisher={Wiley Online Library}
}

@article{gretton2012kernel,
  title={A kernel two-sample test},
  author={Gretton, Arthur and Borgwardt, Karsten M and Rasch, Malte J and Sch{\"o}lkopf, Bernhard and Smola, Alexander},
  journal={The Journal of Machine Learning Research},
  volume={13},
  number={1},
  pages={723--773},
  year={2012},
  publisher={JMLR. org}
}

@inproceedings{berndt1994dtw,
  title={Using dynamic time warping to find patterns in time series},
  author={Berndt, Donald J and Clifford, James},
  booktitle={Proceedings of the 3rd international conference on knowledge discovery and data mining},
  pages={359--370},
  year={1994}
}

@inproceedings{norgaard2018synthetic,
  title={Synthetic sensor data generation for health applications: A supervised deep learning approach},
  author={Norgaard, Skyler and Saeedi, Ramyar and Sasani, Keyvan and Gebremedhin, Assefaw H},
  booktitle={2018 40th Annual International Conference of the IEEE Engineering in Medicine and Biology Society (EMBC)},
  pages={1164--1167},
  year={2018},
  organization={IEEE}
}

@inproceedings{crowston2012amazon,
  title={Amazon mechanical turk: A research tool for organizations and information systems scholars},
  author={Crowston, Kevin},
  booktitle={Shaping the Future of ICT Research. Methods and Approaches: IFIP WG 8.2, Working Conference, Tampa, FL, USA, December 13-14, 2012. Proceedings},
  pages={210--221},
  year={2012},
  organization={Springer}
}

@inproceedings{
bao2022why,
title={Why Are Conditional Generative Models Better Than Unconditional Ones?},
author={Fan Bao and Chongxuan Li and Jiacheng Sun and Jun Zhu},
booktitle={NeurIPS 2022 Workshop on Score-Based Methods},
year={2022}
}

@inproceedings{leng2023generating,
  title={Generating virtual on-body accelerometer data from virtual textual descriptions for human activity recognition},
  author={Leng, Zikang and Kwon, Hyeokhyen and Pl{\"o}tz, Thomas},
  booktitle={Proceedings of the 2023 ACM International Symposium on Wearable Computers},
  pages={39--43},
  year={2023}
}

@article{katada2022effects,
  title={Effects of physiological signals in different types of multimodal sentiment estimation},
  author={Katada, Shun and Okada, Shogo and Komatani, Kazunori},
  journal={IEEE Transactions on Affective Computing},
  volume={14},
  number={3},
  pages={2443--2457},
  year={2022},
  publisher={IEEE}
}

@article{gao2024explainable,
  title={An explainable longitudinal multi-modal fusion model for predicting neoadjuvant therapy response in women with breast cancer},
  author={Gao, Yuan and Ventura-Diaz, Sofia and Wang, Xin and He, Muzhen and Xu, Zeyan and Weir, Arlene and Zhou, Hong-Yu and Zhang, Tianyu and van Duijnhoven, Frederieke H and Han, Luyi and others},
  journal={Nature Communications},
  volume={15},
  number={1},
  pages={9613},
  year={2024},
  publisher={Nature Publishing Group UK London}
}

@article{triantafyllidis2022deep,
  title={Deep learning in mHealth for cardiovascular disease, diabetes, and cancer: systematic review},
  author={Triantafyllidis, Andreas and Kondylakis, Haridimos and Katehakis, Dimitrios and Kouroubali, Angelina and Koumakis, Lefteris and Marias, Kostas and Alexiadis, Anastasios and Votis, Konstantinos and Tzovaras, Dimitrios and others},
  journal={JMIR mHealth and uHealth},
  volume={10},
  number={4},
  pages={e32344},
  year={2022},
  publisher={JMIR Publications Inc., Toronto, Canada}
}

@article{junaid2023explainable,
  title={Explainable machine learning models based on multimodal time-series data for the early detection of Parkinson’s disease},
  author={Junaid, Muhammad and Ali, Sajid and Eid, Fatma and El-Sappagh, Shaker and Abuhmed, Tamer},
  journal={Computer Methods and Programs in Biomedicine},
  volume={234},
  pages={107495},
  year={2023},
  publisher={Elsevier}
}

@article{yang2023ts,
  title={Ts-gan: Time-series gan for sensor-based health data augmentation},
  author={Yang, Zhenyu and Li, Yantao and Zhou, Gang},
  journal={ACM Transactions on Computing for Healthcare},
  volume={4},
  number={2},
  pages={1--21},
  year={2023},
  publisher={ACM New York, NY}
}

@article{wijesinghe2024ps,
  title={PS-FedGAN: An efficient federated learning framework with strong data privacy},
  author={Wijesinghe, Achintha and Zhang, Songyang and Ding, Zhi},
  journal={IEEE Internet of Things Journal},
  volume={11},
  number={16},
  pages={27584--27596},
  year={2024},
  publisher={IEEE}
}

@inproceedings{pennisi2023privacy,
  title={A privacy-preserving walk in the latent space of generative models for medical applications},
  author={Pennisi, Matteo and Proietto Salanitri, Federica and Bellitto, Giovanni and Palazzo, Simone and Bagci, Ulas and Spampinato, Concetto},
  booktitle={International Conference on Medical Image Computing and Computer-Assisted Intervention},
  pages={422--431},
  year={2023},
  organization={Springer}
}

@inproceedings{
chen2024pathformer,
title={Pathformer: Multi-scale Transformers with Adaptive Pathways for Time Series Forecasting},
author={Peng Chen and Yingying ZHANG and Yunyao Cheng and Yang Shu and Yihang Wang and Qingsong Wen and Bin Yang and Chenjuan Guo},
booktitle={The Twelfth International Conference on Learning Representations},
year={2024}
}

@inproceedings{
liu2025timerxl,
title={Timer-{XL}: Long-Context Transformers for Unified Time Series Forecasting},
author={Yong Liu and Guo Qin and Xiangdong Huang and Jianmin Wang and Mingsheng Long},
booktitle={The Thirteenth International Conference on Learning Representations},
year={2025}
}

@inproceedings{
gu2025verbalts,
title={Verbal{TS}: Generating Time Series from Texts},
author={Shuqi Gu and Chuyue Li and Baoyu Jing and Kan Ren},
booktitle={Forty-second International Conference on Machine Learning},
year={2025}
}

@article{yu2024human,
  title={Human activity recognition using deep residual convolutional network based on wearable sensors},
  author={Yu, Xugao and Al-Qaness, Mohammed AA},
  journal={IEEE Journal of Biomedical and Health Informatics},
  year={2024},
  publisher={IEEE}
}

@inproceedings{zhang2023crossformer,
  title={Crossformer: Transformer utilizing cross-dimension dependency for multivariate time series forecasting},
  author={Zhang, Yunhao and Yan, Junchi},
  booktitle={The eleventh international conference on learning representations},
  year={2023}
}

\end{document}